\documentclass{article}

\usepackage{microtype}
\usepackage{graphicx}
\usepackage{subfigure}
\usepackage{booktabs} 

\usepackage{hyperref}

\usepackage[accepted]{icml2023}

\usepackage{amsmath}
\usepackage{amssymb}
\usepackage{mathtools}
\usepackage{booktabs}    
\usepackage{amsthm}
\usepackage{bbm}
\newcommand{\indep}{\small {\raisebox{0.05em}{\rotatebox[origin=c]{90}{$\models$}}}}
\usepackage{subfigure}

\usepackage[capitalize,noabbrev]{cleveref}
\usepackage{enumitem}
\setlist[itemize]{noitemsep, nolistsep}
\setlist[enumerate]{noitemsep, nolistsep}
\theoremstyle{plain}
\newtheorem{theorem}{Theorem}[section]

\theoremstyle{definition}

\newtheorem{assumption}[theorem]{Assumption}
\theoremstyle{remark}

\usepackage[textsize=tiny]{todonotes}
\newcommand{\squish}[1]{{#1\parfillskip=0pt\par}}

\icmltitlerunning{Towards Demystification of the Model Selection Dilemma in Heterogeneous Treatment Effect Estimation}

\begin{document}

\twocolumn[
\icmltitle{In Search of Insights, Not Magic Bullets: Towards Demystification of the Model Selection Dilemma in Heterogeneous Treatment Effect Estimation} 



\icmlsetsymbol{equal}{*}

\begin{icmlauthorlist}
\icmlauthor{Alicia Curth}{yyy}
\icmlauthor{Mihaela van der Schaar}{yyy,comp}
\end{icmlauthorlist}

\icmlaffiliation{yyy}{Department of Applied Mathematics and Theoretical Physics, University of Cambridge, UK}
\icmlaffiliation{comp}{The Alan Turing Institute}

\icmlcorrespondingauthor{Alicia Curth}{amc253@cam.ac.uk}

\icmlkeywords{Heterogeneous Treatment Effect, CATE, ITE, Model Selection}

\vskip 0.3in
]



\printAffiliationsAndNotice{} 

\begin{abstract}
Personalized treatment effect estimates are often of interest in high-stakes applications -- thus, before deploying a model estimating such effects in practice, one needs to be sure that the best candidate from the ever-growing machine learning toolbox for this task was chosen. Unfortunately, due to the absence of counterfactual information in practice, it is usually not possible to rely on standard validation metrics for doing so, leading to a well-known model selection dilemma in the treatment effect estimation literature. While some solutions have recently been investigated, systematic understanding of the strengths and weaknesses of different model selection criteria is still lacking. In this paper, instead of attempting to declare a global `winner', we therefore empirically investigate success- \textit{and} failure modes of different selection criteria. We highlight that there is a complex interplay between selection strategies, candidate estimators and the data used for comparing them, and provide interesting insights into the relative (dis)advantages of different criteria alongside desiderata for the design of further illuminating empirical studies in this context.

\end{abstract}

\section{Introduction}\label{introduction}
Applications in which the causal effects of treatments (or  actions, interventions \& policies) are of interest  are ubiquitous in empirical science, and personalized effect estimates could ultimately be used to improve decision making in many domains, including healthcare, economics and marketing. Machine learning (ML) has shown great promise in providing such personalized effect estimates \citep{bica2021real}, and the ML literature on the topic has matured over the last five years: a plethora of new methods for estimating conditional average treatment effects (CATE) have been proposed recently, including \textit{method-agnostic} so-called meta-learner strategies that could be implemented using any ML prediction method \citep{kunzel2019metalearners, nie2021quasi, kennedy2020optimal, curth2021nonparametric} as well as adaptations of specific ML methods \citep{shalit2017estimating, alaa2018limits, wager2018estimation} to the treatment effect estimation context. 

Personalized treatment effect estimates, crucially, are often of interest in safety-critical applications, particularly in medicine and policy making. Thus, prior to use in practice we would like to ensure that we have selected a ML estimator from this vast toolbox that is \textit{trustworthy} and outputs the best possible estimates. While this sounds like a straightforward requirement, it remains a big hurdle in practice: because of the \textit{fundamental problem of causal inference} \cite{holland1986statistics}, ground truth treatment effects are usually not available to perform standard model validation, and alternative solutions need to be considered to overcome this \textit{model selection dilemma}. As we discuss below, despite recent proposals of new model selection strategies and recent empirical studies comparing different strategies, we believe that there is still a general lack of understanding of the (dis)advantages of different strategies and how they are entangled with underlying data-generating processes (DGPs) -- which we aim to provide in this work.

\textbf{Related work.} Despite its practical relevance, the problem of model selection for heterogeneous treatment effect estimation has received only very limited attention so far (this stands also in stark contrast to the plethora of new estimators proposed in recent years, see Appendix \ref{app:lit} for an overview). An intuitive and often applied solution is to rely on a simple prediction-type validation and evaluate a model's performance in \textit{predicting the observed (factual) outcome} associated with the factual treatment \textit{instead} of evaluating the quality of the \textit{effect estimate}. More targeted alternatives have recently been developed: \citet{rolling2014model} propose to construct approximate effect validation targets by matching the nearest treated and control units and comparing their outcomes, \citet{nie2021quasi} highlight that their R-learner objective could also be used for model selection, \citet{saito2020counterfactual} similarly propose the use of a criterion that corresponds to \citet{kennedy2020optimal}'s meta-learner objective and \citet{alaa2019validating} propose to rely on influence functions to de-biase plugin estimates. We are aware of two independent benchmarking studies that compare (subsets of) such criteria: \citet{schuler2018comparison} find that the R-learner objective performs best overall, while \citet{mahajan2022empirical} find that no criterion dominates all others over all datasets considered (and in particular do not find the R-learner objective to perform remarkably), which highlights to us that there is much room for understanding of the relative strengths but also relative weaknesses of different selection criteria.

 \textbf{Contributions.} In this paper, we focus on building \textit{systematic understanding} of the (dis)advantages of different model selection criteria.  Note that therefore our aim is not to propose new methodology, but rather to establish understanding and insight into the tools already available in the literature. We believe that this is one of the most crucial and necessary next steps for this community in order to enable actual adoption of personalized treatment effect estimators in practice, and may inspire further methodological research to fill gaps highlighted by this understanding. In doing so, we make three contributions:\vspace{-.25cm}
 \begin{enumerate}[noitemsep, leftmargin=*]
     \item We develop intuition for a highly complex selection problem: we shine light on its inherent challenges, provide structure to existing work by presenting a classification of existing criteria, and use these insights to derive hypotheses for their relative performance.
     \item We present desiderata for experimental design that enable us to disentangle the complex forces at play in this problem: we advocate for better experiments that allow to systematically investigate the interplay between DGP, candidate estimators and selection criteria through reliance on data-generating processes with interesting axes of variation and more transparent reporting practices.
     \item We provide new insights into the CATE model selection problem through an empirical investigation of the success and failure modes of existing criteria, and conclude that no existing selection criterion is globally best across all experimental conditions we consider. Next to highlighting some performance trends across the different types of selection criteria, we mainly focus on investigating i) congeniality biases between candidate estimators and selection criteria imbued with similar inductive biases in their construction and ii) what factual selection criteria can(not) achieve. We find that i) selection criteria relying on plug-in estimates of treatment effects are likely to favor estimators that resemble their plug-ins, while in selection criteria relying on pseudo-outcomes such congeniality biases are less pronounced, and that ii) factual selection sometimes underperforms not only because it cannot evaluate all types of CATE estimators, but also because it is not well-targeted at effect estimates.
 \end{enumerate}

\section{Problem Setting}
We consider the by now standard CATE estimation setup within the potential outcomes framework \cite{rubin2005causal}. That is, we assume access to a dataset consisting of $n$ i.i.d. tuples $(X, A, Y)$, where $Y$ is an outcome of interest, $X$ consists of \textit{pre-treatment} covariates and  $A \in \{0, 1\}$ is a binary treatment (action, intervention or policy), which is assigned according to some (often unknown) propensity $\pi(x)=\mathbb{P}(A=1|X=x)$. We assume that conceptually each individual is a priori associated with two \textit{potential outcomes (POs)} $Y(a), a \in \{0, 1\}$, capturing outcome under either treatment $a$, however, we observe only the outcome associated to the treatment $A$ actually received, i.e. $Y=Y(A)$, We can thus naturally define an individualized treatment effect through the (unobserved) PO contrast $Y(1)\!-\!Y(0)$. We focus on estimating the conditional average treatment effect (CATE) $\tau(x)$, i.e. the expected PO difference for an individual with covariates $X=x$:
\begin{equation}
    \tau(x) = \mathbb{E}[Y(1)-Y(0)|X=x] = \mu_1(x) - \mu_0(x)
\end{equation}
where $\mu_a(x)=\mathbb{E}[Y(a)|X=x]$. To ensure that effects are identifiable and nonparametrically estimable from observational data, we rely on the standard ignorability assumptions \cite{rosenbaum1983central}:
\begin{assumption}[\textbf{Ignorability.}] (i) \textit{Consistency.} For an individual with treatment assignment $A$, we observe the associated potential outcome, i.e. $Y=Y(A)$. (ii) \textit{Unconfoundedness.} There are no unobserved confounders, so that $Y(0), Y(1) \indep A | X$. (iii) \textit{Overlap.} Treatment assignment is non-deterministic, i.e. $\pi(x) \in (0, 1)$.
\end{assumption}
Then, $\mathbb{E}[Y(a)|X\!=\!x]=\mathbb{E}[Y|X\!=\!x, A\!=\!a]$ so that observed statistical associations have a causal interpretation.

\subsection{CATE Estimation Strategies}\label{sec:estimators}
A plethora of strategies for estimating CATE have been proposed in the recent literature. One strand of this work has recently relied on the \textit{meta-learner} framework of \citet{kunzel2019metalearners}, where a meta-learner provides a `recipe' for estimating CATE using \textit{any arbitrary}\footnote{as opposed to \textit{adaptation of specific} ML methods proposed in another stream of work, as discussed further in Appendix \ref{app:lit}.} ML method $\mathcal{M}$. Due to their ease of implementation with different underlying ML methods, existing theoretical understanding and correspondence to the model selection strategies discussed in the following section, we focus on the problem of choosing between such meta-learners in this paper.

Following \citet{curth2021nonparametric} we distinguish between (i) \textit{indirect estimation strategies}, which estimate CATE \textit{indirectly} by outputting estimates $\hat{\mu}_a(x)$ of the PO regressions and then setting $\hat{\tau}=\hat{\mu}_1(x)-\hat{\mu}_0(x)$,  and (ii) \textit{direct estimation strategies} which output an estimate $\hat{\tau}(x)$ \textit{directly} without outputting PO estimates as byproducts. 
\citet{kunzel2019metalearners} discuss two \textit{indirect learners}: a T-learner strategy, where the training data is split by treatment group and $\mathcal{M}$ is trained independently (twice) on each sample to output \textbf{t}wo regressors $\{\hat{\mu}_0(x), \hat{\mu}_1(x)\}$, and a S-learner strategy, where the treatment indicator $A$ is simply appended to $X$ so that $\mathcal{M}$ can be trained a \textbf{s}ingle time using covariates $(X, A)$ and outputting a single estimated function $\hat{\mu}(x, a)$ that can be used to impute both POs. Because the latter formulation can lead to implicit regularization of CATE \cite{schuler2018comparison} (as any heterogeneous effect has to be represented by \textit{learned interaction terms} of $X$ and $A$), we also include an extended version (ES-learner) which is trained on the covariates $(X, X*A, A)$ explicitly.

Alternatively, there exist \textit{direct estimators} that output an estimate of the CATE \textit{only} by relying on estimates of (some of) the nuisance parameters $\eta = (\mu_0(x), \mu_1(x), \pi(x))$ obtained in a pre-processing step using ML method $\mathcal{M}$. Most such strategies, in particular, X-learner \cite{kunzel2019metalearners}, the DR-learner \cite{kennedy2020optimal} and RA- and PW-learner \cite{curth2021nonparametric}, rely on a pseudo-outcome approach where, using plug-in nuisance estimates $\hat{\eta}$, one constructs a pseudo outcome for which it holds that $\mathbb{E}[Y_{\eta}|X=x]=\tau(x)$ for ground truth nuisance parameters ${\eta}$, and then regresses $Y_{\hat{\eta}}$ on $X$ using $\mathcal{M}$ to give an estimate $\hat{\tau}(x)$. \citet{nie2021quasi}'s R-learner is similar in spirit but relies on a modified loss function instead. These multi-stage estimation procedures have recently gained popularity in the literature because they have good theoretical properties \cite{curth2021nonparametric}, are more robust \cite{kennedy2020optimal} and have been observed to perform better across a variety of scenarios than vanilla S- and T-learner individually \cite{nie2021quasi}.

\section{CATE Model Selection: Understanding Challenges and Existing Strategies}
We study the problem of selecting an estimator from a set of  CATE estimators $\mathcal{T}\!=\!\{\hat{\tau}_1(\cdot), \ldots, \hat{\tau}_K(\cdot)\}$, containing different meta-learner+ML method combinations, that minimizes the precision of estimating heterogeneous effects (PEHE) \cite{hill2011bayesian}, the root-mean-squared error of estimating the underlying effect function over a test-set of size $n$:
\begin{equation*}
\arg \min_{\hat{\tau}_k \in \mathcal{T}} \mathcal{E}^{oracle}_{\tau}(\hat{\tau}_k)=\sqrt{\frac{1}{n}\sum^{n}_{i=1}(\tau(X_i) - \hat{\tau}_k(X_i))^2}
\end{equation*}
This metric is an \textit{oracle} metric because it cannot be evaluated in practice -- making the CATE model selection problem highly nontrivial. Below, we therefore provide an in-depth discussion of the inherent challenges of this problem, and then establish a classification of model selection criteria that have been proposed to overcome these challenges. 

\subsection{What makes CATE model selection challenging?}
\squish{\textbf{Challenge 1: Lack of supervised signal for the individual treatment effect. } Due to the \textit{fundamental problem of causal inference} \cite{holland1986statistics}, i.e. the fact that we can only \textit{either} observe $Y(0)$ or $Y(1)$ for any one individual, the true supervised target label $Y(1)-Y(0)$ for estimation of $\mathbb{E}[Y(1)-Y(0)|X=x]=\tau(x)$ is not available for model selection through a standard held-out validation approach. This lack of supervised label is also the issue that motivated the construction of the direct meta-learners for \textit{estimation} of effects using pseudo-outcomes \cite{kennedy2020optimal, curth2021nonparametric}, but it results in another challenge for model selection: even though direct estimation of effects is possible, it is not possible to compare multiple direct estimators on basis of their output $\hat{\tau}_k(x)$ using \textit{observed data only} because there is no natural outcome to validate $\hat{\tau}_k(x)$ against directly.}

\textbf{Challenge 2: Confounding leads to covariate shift between treatment groups. } One straightforward option to validate estimators using factual (observed) outcomes only is to simply evaluate them based on their outcome prediction ability; i.e. to use $Y_i - \hat{\mu}_{A_i}(X_i)$ for validation. This, however, is only an option for evaluating indirect learners because it requires an output $\hat{\mu}_a(x)$. Even when one is willing to restrict attention only to indirect learners to make use of a factual evaluation strategy, a remaining inherent challenge is that evaluating $\hat{\mu}_a(x)$ only on individuals with $A=a$ observed inherently suffers from \textit{covariate shift} whenever $\pi(x)$ is not constant because treatment is not assigned completely randomly. This too is a challenge also when estimating effects \cite{shalit2017estimating}.

\textbf{Challenge 3: Selection for good PO estimation and CATE estimation may not be the same.} Finally, even when selecting only among indirect learners and in absence of covariate shift, in finite samples the estimator with the best performance on estimating (potential) outcomes might not do best at estimating CATE. It is clear that when models are correctly specified and unlimited data is available, perfectly estimating the POs will immediately lead to perfect CATE estimates. However, when data is limited and/or the model is misspecified, this might lead to a trade-off between estimating the POs well and estimating their difference\footnote{To see this, note that when comparing an indirect estimator $\hat{\mu}^1_a(x)=\{\mu_1(x) + \tilde{\epsilon}_1(x), \mu_1(x) - \tilde{\epsilon}_1(x)\}$ to an estimator $\hat{\mu}^2_a(x)=\{\mu_1(x) + \tilde{\epsilon}_2(x), \mu_1(x) + \tilde{\epsilon}_2(x)\}$ with estimation errors $\tilde{\epsilon}_1(x) < \tilde{\epsilon}_2(x)$ for all $x$, the MSE of estimating the POs will be lower for estimator $\hat{\mu}^1_a(x)$ -- yet its estimation error on CATE will be $2 \tilde{\epsilon}_1(x)$ for every $x$ while estimator $\hat{\mu}^2_a(x)$ with worse estimation on the POs will have CATE estimation error $0$. }.
\subsection{Categorizing model selection criteria}
\begin{figure}[!t]
    \centering
    \includegraphics[width=0.99\columnwidth]{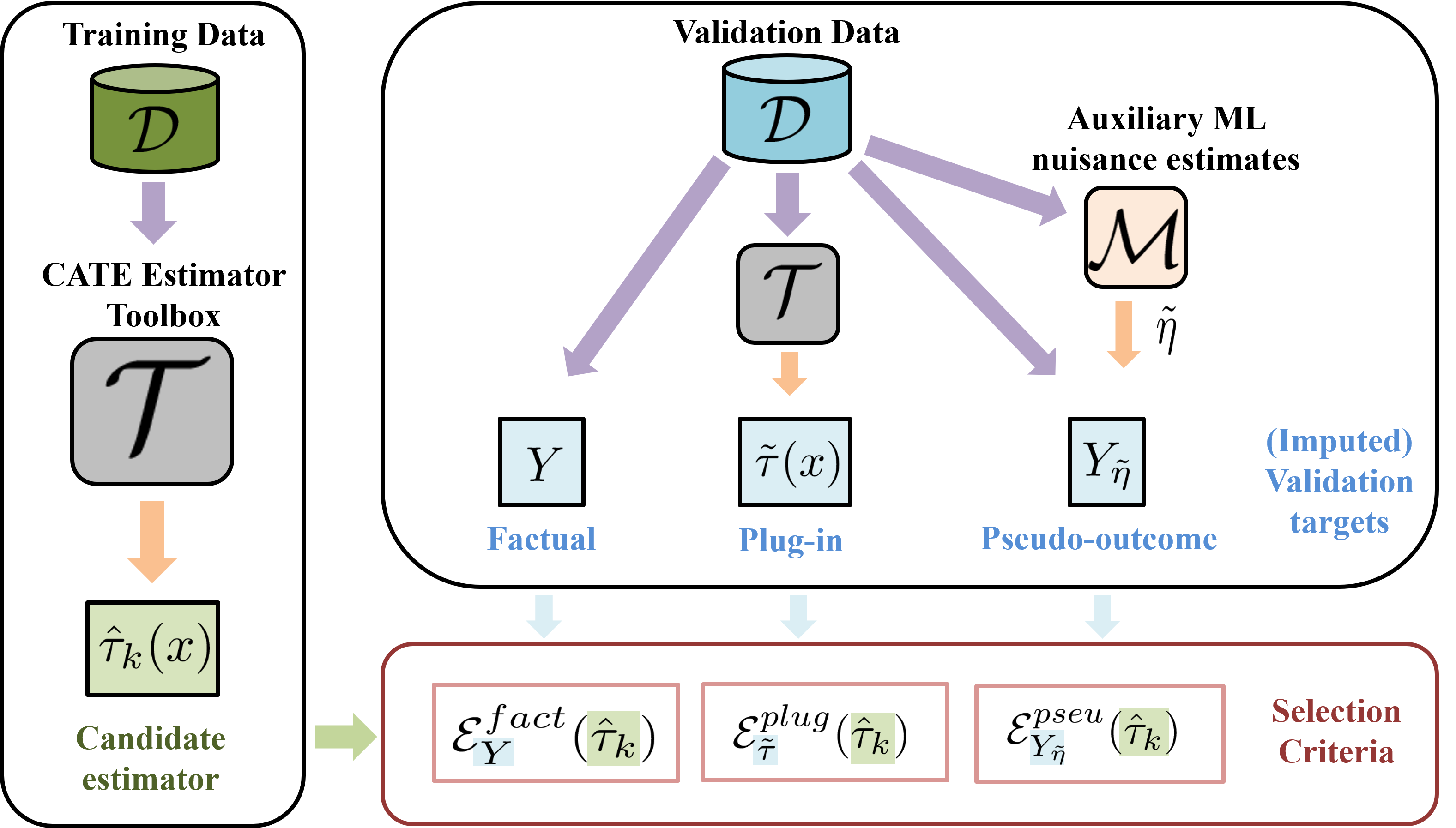}\vspace{-10pt}
    \caption{Conceptual overview of the considered selection criteria.}
    \label{fig:overviewfig}\vspace{-5pt}
\end{figure}

To overcome the challenges discussed above, numerous alternative model selection criteria have been used or proposed in related work. Below, we establish a classification of existing model selection criteria into three categories based on their most salient characteristics: we consider factual (prediction) criteria, plug-in surrogate criteria and pseudo-outcome surrogate criteria. We provide a conceptual overview of the three types of strategies in Fig. \ref{fig:overviewfig}. Overall, we consider a similar set of model selection criteria as the (union of) the benchmarking studies presented in \citet{schuler2018comparison, mahajan2022empirical}\footnote{\citet{schuler2018comparison}, predating publication of most model selection papers, miss some of the plug-in and pseudo-outcome criteria. \citet{mahajan2022empirical} only do not consider factual (prediction) criteria, which we consider of major interest as discussed in the following section. We only drop \textit{policy value} criteria, i.e. those optimizing the derived treatment policy $\mathbf{1}\{\hat{\tau}_k(x)>0\}$, from consideration, both because we focus on PEHE and because these criteria were shown to underperform \textit{even when evaluated in terms of policy value} \cite{schuler2018comparison}.}.

\textbf{1. Factual (prediction) criteria.} First, as discussed in the previous section, a possible way of evaluating models $\hat{\tau}_k(x)$ that also output a pair of regressors $\{\hat{\mu}^k_0(x), \hat{\mu}^k_1(x)\}$ is to rely on a simple prediction loss considering only the observed potential outcome
\begin{equation*}
 \mathcal{E}^{fact}_{Y}(\hat{\tau}_k)=\sqrt{\frac{1}{n}\sum^{n}_{i=1}(Y_i - \hat{\mu}^k_{A_i}(X_i))^2}
\end{equation*}
In order to correct for possible covariate shift, this can also be transformed into an importance weighted criterion $\mathcal{E}^{fact, w}_{Y}$ using a propensity score estimate in $w(X_i, A_i) = A_i(\hat{\pi}(X_i))^{-1}+(1-A_i)(1-\hat{\pi}(X_i))^{-1}$.

\textbf{2. Plug-in surrogate criteria.} To actually evaluate estimates $\hat{\tau}_k(x)$ \textit{directly} one is thus forced to construct surrogates for CATE. One way of doing so is by fitting a new CATE estimator $\tilde{\tau}(x)$ on held-out data and using this to compare against the estimates:
\begin{equation*}
\mathcal{E}^{plug}_{\tilde{\tau}}(\hat{\tau}_k)=\sqrt{\frac{1}{n}\sum^{n}_{i=1}(\tilde{\tau}(X_i) - \hat{\tau}_k(X_i))^2}
\end{equation*}
For implementation of this criterion, any CATE estimator could be used -- thus trying to select the best plug-in estimator may potentially lead us in a circle and back to the problem we are originally trying to solve. Related work \cite{alaa2019validating, mahajan2022empirical} considered only indirect estimators (S and T-learners) as a plug-in surrogate criterion -- possibly because it is possible to choose between those factually -- but we note that technically any estimator, including direct ones, could be used as $\tilde{\tau}(x)$. 

\textbf{3. Pseudo-outcome surrogate criteria.} Finally, one could make use of the same pseudo-outcome based strategy that underlies the direct learners: given auxiliary nuisance estimates $\tilde{\eta} = (\tilde{\mu}_0(x), \tilde{\mu}_1(x), \tilde{\pi}(x))$ obtained from the validation data using ML method $\mathcal{M}$, one can construct pseudo-outcomes $Y_{\tilde{\eta}}$ for which it holds that for ground truth nuisance parameter $\eta$, $\mathbb{E}[Y_{\eta}|X=x]=\tau(x)$ and -- instead of using them as regression outcomes as in the learners themselves -- one can use them as validation targets in
\begin{equation*}
\mathcal{E}^{pseu}_{Y_{\tilde{\eta}}}(\hat{\tau}_k)=\sqrt{\frac{1}{n}\sum^{n}_{i=1}(Y_{\tilde{\eta}} - \hat{\tau}_k(X_i))^2}
\end{equation*}
which is reasonable because the conditional mean $\mathbb{E}[Y_{\tilde{\eta}}|X=x]$ minimizes the MSE $\mathcal{E}^{pseu}_{Y_{\tilde{\eta}}}(\cdot)^2$ in expectation.
Use of the doubly robust pseudo-outcome, 
\begin{equation*}
\begin{split}
{Y}_{DR, \tilde{\eta}} = \left(\frac{A}{\hat{\pi}(X)}- \frac{(1-A)}{1-\hat{\pi}(X)}\right) Y + \\ \left[\left(1 - \frac{A}{\hat{\pi}(X)}\right) \hat{\mu}_1(x)-\left(1 - \frac{1-A}{1-\hat{\pi}(X)}\right)\hat{\mu}_0(X)\right]
\end{split}
\end{equation*} 
which is what the direct meta-learner known as the DR-learner \cite{kennedy2020optimal} is based on, gives rise to the selection criterion proposed in \citet{saito2020counterfactual}. We note here that it would also be possible to use the other meta-learner pseudo-outcomes discussed in \citet{curth2021nonparametric} for this purpose, e.g. the singly-robust propensity-weighted $\textstyle{{Y}_{PW, \tilde{\eta}} = \left(\frac{A}{\hat{\pi}(X)}- \frac{(1-A)}{1-\hat{\pi}(X)}\right) Y}$. We also consider \citet{rolling2014model}'s matching based model selection strategy to fall into this category. In this case $
 \textstyle{Y_{\text{match}, \tilde{\eta}}=(2A_i-1)(Y_i - \tilde{NN}_{1-A_i}(X_i))}$ where $\tilde{NN}_a(X)$  is the nearest neighbor of $X$ in treatment group $a$; this essentially corresponds to the pseudo-outcome associated with the RA-learner of \citet{curth2021nonparametric}, implemented using 1-NN regression to estimate the nuisance parameters $\hat{\mu}_a(x)$. We also put \citet{alaa2019validating}'s influence function based criterion, which we discuss in Appendix \ref{app:details}, into this category. Finally, the R-learner objective of \citet{nie2021quasi}, which requires an estimate of the treatment-unconditional mean $\mu(x)=\mathbb{E}[Y|X=x]$, relies on a similar idea\footnote{$\mathcal{E}^{pseu}_{R}$ can be rewritten in pseudo-outcome form  as  $\textstyle{{Y}_{R, \tilde{\eta}} = \frac{Y_i - \tilde{\mu}(X_i)}{A_i - \tilde{\pi}(X_i)}}$ if combined with weights $(A_i - \tilde{\pi}(X_i))^2$ to be used in the sum inside the RMSE \citep{knaus2021machine}.} and can also been used for the selection task \cite{nie2021quasi}, resulting in the criterion
\begin{equation*}
\mathcal{E}^{pseu}_{R}(\hat{\tau}_k)=\sqrt{\frac{1}{n}\sum^{n}_{i=1}(Y_i \!-\! \tilde{\mu}(X_i)\!-\! \hat{\tau}_k(X_i)(A_i\! -\! \tilde{\pi}(X_i)))^2}
\end{equation*}

\section{Demystifying the Model Selection Dilemma}\label{sec:demyst}
Having established a high-level classification of model selection strategies, we can now build intuition and hypothesize about their expected (dis)advantages based on their inherent characteristics  (Sec. \ref{sec:expectations}). These expectations lead to numerous interesting research questions, which we believe can only be disentangled by designing carefully controlled experiments -- a feature that we would argue related work has neglected so far. Therefore, we then move to discuss design principles for construction of empirical studies of the CATE model selection problem (Sec. \ref{sec:expdesign}), which we will then apply in our experiment section. 

\subsection{Comparing model selection criteria: expectations on advantages and disadvantages}\label{sec:expectations}
Combining on our high-level overview of different strategies for CATE model selection with the previously discussed challenges, we argue that every \textit{class} of criteria comes with their own inherent (dis)advantages. 

 \textbf{Factual (prediction) criteria.} We believe that factual criteria could be very appealing for use in practice because -- at least the unweighted $\mathcal{E}^{fact}_{Y}$ -- does not require estimates of any nuisance parameters and \textit{only relies on observed data}; this means that there is no additional overhead and its results could be considered trustworthy in the sense that there is no dependence on possibly misspecified or biased nuisance estimates. However, such criteria mainly evaluate the performance in terms of estimation of the POs, which, as discussed above, may wrongly prioritize good fit on the POs over good CATE fit -- while the latter cannot be measured at all. This last point, crucially, also means that the factual criterion  \textit{cannot} evaluate all types of methods, excluding all direct estimators in particular, and therefore has to select among a smaller set $\mathcal{T}_{indirect}\subset \mathcal{T}$ of all possible estimators -- potentially missing out on the estimators with the best performance by construction. 

 \squish{\textbf{Plug-in surrogate criteria.} Plug-in surrogate criteria have the clear advantage over factual criteria that they can evaluate \textit{all} types of estimators and are targeted at the outcome of interest (i.e. the CATE). Yet, because a plug-in estimate $\tilde{\tau}(x)$ is needed, this introduces additional potential for estimation error or variance. Further, $\tilde{\tau}(x)$ could be any CATE estimator, thus choosing a good plugin $\tilde{\tau}(x)$ leads us back to the dilemma we were initially trying to overcome. Finally, we believe that such surrogate criteria may also suffer from a phenomenon we will refer to as \textbf{\textit{congeniality bias}}: they may advantage CATE estimators $\hat{\tau}_k(x)$ that are \textit{structurally similar} to their plug-in estimator $\tilde{\tau}(x)$. Even though $\hat{\tau}_k(x)$ and $\tilde{\tau}(x)$ should be fit on different data folds, we expect that a plug-in criterion $\tilde{\tau}(x)$ may still prefer estimators imbued with similar inductive biases. That is, we expect that e.g. a criterion using plug-in surrogate $\tilde{\tau}(x)$ implemented using linear regression may favor CATE estimators $\hat{\tau}_k(x)$ similarly relying on linear regressions (over estimators implemented using other methods $\mathcal{M}$), and one relying on an S-learner surrogate $\tilde{\tau}(x)$ may have a preference for selecting S-learners $\hat{\tau}_k(x)$. Here, we borrow the term `congeniality bias’ from the psychology literature, where it is used to indicate that individuals may have a systematic preference for information consistent with current beliefs \citep{hart2009feeling}.}

 \textbf{Pseudo-outcome surrogate criteria.} We expect these criteria to share the advantages, and some of the disadvantages of the plug-in criteria (namely the need to estimate additional parameters, and resulting possibility for increased error or variance). Because they do not use a final $\tau(x)$ estimate but only a pseudo-outcome $Y_{\tilde{\eta}}$, we expect that they might be \textit{less} likely to suffer from congeniality bias, but could still favor estimators with similar inductive biases, e.g. direct estimators trained on the same pseudo-outcome. 

\subsubsection{Resulting research questions.} 
This paper was motivated by many of the interesting research questions outlined below that naturally follow from the discussion above, none of which we believe have been addressed in related work-- we believe that this is a result of the fact that the focus so far has been on establishing \textit{global best performance} of some criterion. Instead, we are interested in understanding scenarios in which there could be performance differences of competing criteria, in the hope that this will help practitioners in choosing the right criterion in their specific application. In particular, we are interested in exploring three questions in depth in this paper:\vspace{-.3cm}
\begin{itemize}[noitemsep, leftmargin=*]
\item \textbf{Q1.} When do which selection criteria work better or worse? If there are systematic patterns, do they parallel those observed for the different \textit{estimation strategies} in prior work?
    \item \textbf{Q2.} Do surrogate selection criteria truly suffer from congeniality bias as expected? Are there differences between the different types?
    \item \textbf{Q3.} What and when do we lose out on by relying on factual validation $\mathcal{E}^{fact}_{\mu_a}$? Does it matter that we restrict the estimator pool? Does it matter that we are optimizing for the wrong target?
\end{itemize}

\subsection{Establishing desiderata for experimental design}\label{sec:expdesign}
As outlined above, we believe that there are many very interesting questions to explore in this CATE model selection dilemma. Existing empirical studies that we are aware of -- both those proposing new model selection criteria \cite{alaa2019validating, saito2020counterfactual} and those benchmarking existing ones \cite{schuler2018comparison, mahajan2022empirical} -- have mainly focused on the question \textit{`what strategy works best globally?'}, and have taken a black-box approach in doing so: by considering opaque datasets for benchmarking, by considering a large inextricable set of estimators to select from and by mainly reporting on averages across a number of different DGPs. 

It has recently been highlighted that the benchmarking practices in the ML CATE estimation literature more generally have many shortcomings \cite{curth2021really}, especially the reliance on single semi-synthetic datasets\footnote{Note that, due to the lack of $Y(1)\!-\!Y(0)$ in real data it is generally \textit{necessary to simulate} outcomes in experiments to allow for known ground truth; most existing benchmark datasets are \textit{semi-synthetic} in that they use real covariates but simulate outcomes.} that encode very specific problem characteristics in their DGP, without discussing the effect of these choices. We therefore believe that it is crucial to carefully design controlled experimental environments that allow to disambiguate the effects that different components of a DGP may have on selection criterion performance. Below we discuss desiderata for designing an empirical study that allows for the insight into model selection performance we seek, which we use to design an empirical study in the following section.

\textbf{1. Use DGPs with interesting `experimental knobs'.} To gain systematic understanding of \textit{when} different selection criteria work well, we need to be able to \textit{systematically} vary the underlying experimental characteristics -- as suggested by \citet{dorie2019automated} we therefore design simulations that enable turning of important `experimental knobs'. In pursuit of interesting insights into relative performance of selection criteria, we will therefore choose to investigate axes that have been shown to matter for estimator performance itself as well as others that we would expect to matter.

{\textbf{2. Examine the performance of candidate estimators in $\mathcal{T}$.}} To understand when different selectors perform well, we believe that it is important to first establish how the underlying candidate estimators in $\mathcal{T}$ perform. All related work that we are aware of skip this step, yet we consider it crucial because the performance of different selection strategies may be deeply entangled with the performance of the underlying estimators: if, for example, there is congeniality between selectors and estimators, and/or a specific type of estimator is advantaged on a specific dataset, then the corresponding selector may perform well by construction.

\textbf{3. Analyze how and \textit{when} performance of selection criteria varies.}  Finally, once DGPs are constructed and estimators examined, we aim to analyze the performance of the selection criteria in detail. That is, we wish to explicitly understand how the relative performance of selectors varies as an experimental knob is turned. In many related works this is not possible as results are reported as \textit{averages} across many different settings \cite{alaa2019validating,  mahajan2022empirical}, obfuscating possible interesting insights into systematic performance differences.

\begin{figure*}[!t]
    \centering
    \includegraphics[width=.99\textwidth]{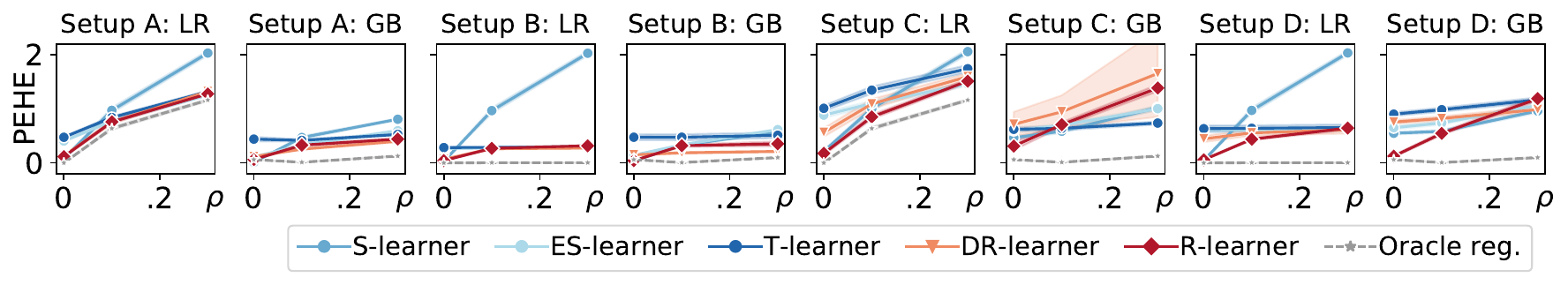}\vspace{-15pt}
    \caption{\textbf{Error in CATE estimation (PEHE) for the different candidate estimators in $\mathcal{T}$.} \small All learners are implemented using linear regressions (LR) and extreme gradient boosting (GB), and considered across 4 different settings where the complexity of $\tau(x)$ increases in $\rho$. Shaded area indicates one SE.}\vspace{-5pt}
    \label{fig:learners}
\end{figure*}
\section{Empirical Study}
\textbf{Setup. } In this section, we conduct an empirical study comparing CATE  selection criteria following the three steps outlined above. Throughout, we rely on two ML-methods $\mathcal{M}$ to instantiate all meta-learners and selection criteria: Extreme Gradient Boosted Trees \cite{chen2016xgboost} (GB) and linear regressions with ridge penalty (LR). We chose these two because they encode very different inductive biases, and allow us to give insights into performance differences between very flexible versus rigid models. Whenever propensity score estimates are needed, we estimate these using logistic regressions. As meta-learners to choose between, we consider (indirect) S-, T- and ES-learners and the (direct) DR- and R-learner. As selection criteria we consider $\mathcal{E}^{fact}_{Y}$ (factual),  $\mathcal{E}^{plug, ES}_{\tilde{\tau}}$, $\mathcal{E}^{plug, T}_{\tilde{\tau}}$, $\mathcal{E}^{plug, DR}_{\tilde{\tau}}$ \& $\mathcal{E}^{plug, R}_{\tilde{\tau}}$ (surrogate plug-in) and $\mathcal{E}^{pseudo, DR}_{Y_{\tilde{\eta}}}$ \& $\mathcal{E}^{pseudo, R}_{Y_{\tilde{\eta}}}$ (surrogate pseudo-outcome) in the main text, further results using criteria that generally performed worse can be found in Appendix \ref{app:others}. Implementation details\footnote{Code to replicate all experiments is available at \texttt{\url{https://github.com/AliciaCurth/CATESelection}}} can be found in Appendix \ref{app:details}.

\begin{figure*}[t]
    \centering
    \includegraphics[width=.99\textwidth]{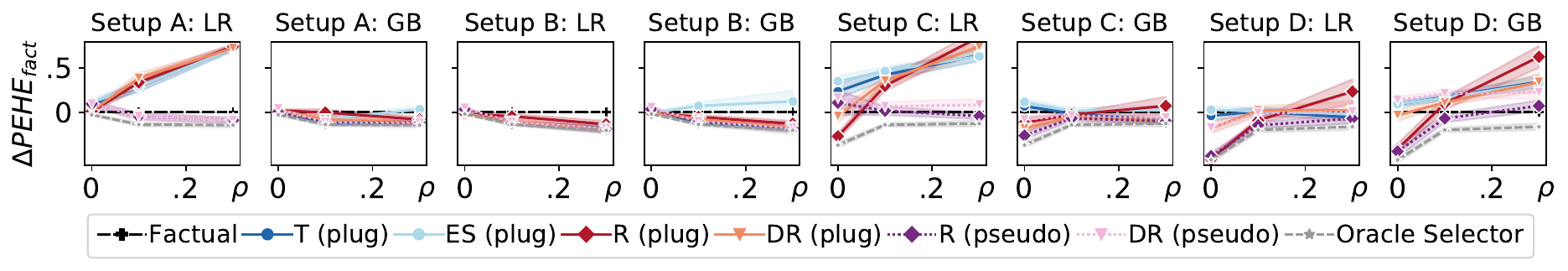}\vspace{-15pt}
    \caption{\textbf{Relative performance of different selection criteria.} \small Plotting $\Delta \text{PEHE}_{fact}$, the difference between the test-PEHE of the factual choice and the model selected by any given selection criterion (lower is better, negative means better than factual) implemented using linear regressions (LR) and extreme gradient boosting (GB). Each criterion gets access to $\mathcal{T}$, i.e. the complete pool of 10 candidate estimators whose performance is shown in Fig. \ref{fig:learners} above. Shaded area indicates one SE.}\vspace{-10pt}
    \label{fig:overallperf}
\end{figure*}
\subsection{Step 1: Designing an insightful DGP.}
We build a DGP loosely inspired by the setup used in \citet{curth2021inductive}: we similarly make use of the covariates $X$ of the ACIC2016 dataset \cite{dorie2019automated}, and simulate our own outcomes and treatment assignments for greater transparency and control. We begin by binarizing all continuous covariates at randomly sampled cutoff points, obtaining processed covariates $X^*$, and then use them in a linear model for $\mu_0(\cdot)$, including up to third-order interaction terms of $X^*$, while $\tau(\cdot)$ is simply linear in $X^*$.  This setup includes 3 main experimental knobs: \vspace{-10pt}
\begin{enumerate}[noitemsep, leftmargin=*]
\item \squish{\textbf{CATE complexity.} Our first experimental knob is $\rho \in \{0, .1, .3\}$, the proportion of non-zero coefficients of inputs in $\tau(\cdot)$, controlling the complexity (sparsity) of $\tau(\cdot)$ -- this was used in \citet{curth2021inductive} and shown to matter for relative performance of \textit{estimators}.}
\item \textbf{Misspecification.}  Second, we introduced a transformation of covariates deliberately because it allows us to explore \textit{the effects of model (mis)specification}. That is, whether algorithms (estimators and selection criteria) are given $X$ or $X^*$ as input $X^{input}$ is the second experimental knob we consider: when given the original data $X$, this implicitly favors tree-based models like GB (this DGP mimics splits in decision trees) because a LR cannot fully recover $X^*$ and hence not learn the patterns in either $\tau(\cdot)$ or $\mu_0(\cdot)$. Even when given the transformed data $X^*$ as  $X^{input}$, a simple LR cannot fit the POs $\mu_a(\cdot)$ due to the higher order interaction terms; however, the treatment effect itself is linear in $X^*$ and could hence be fit with a LR if we observed $Y(1)-Y(0)$.
\item \textbf{Confounding.} Third, we compare confounded  to randomized settings, where only in the former treatments are assigned based on variables that enter $\mu_0(\cdot)$. In the main text, the propensity score logits are linear in $X^{input}$ and can hence be consistently estimated with logistic regressions in all settings. We consider additional settings with other propensities in Appendix \ref{app:otherprop}, where we also consider the effect of imbalance in treatment group sizes (whereas the main text has equal group sizes).
\end{enumerate}\vspace{-5pt}
 {We vary $\rho$ on the x-axis of our plots, which we split into four setups based on characteristics 2 and 3: Setup A is unconfounded and estimators \& selectors get $X$ as input, Setup B is unconfounded and estimators \& selectors get $X^*$ as input, Setup C is confounded and estimators \& selectors get $X$ as input and Setup D is confounded and estimators \& selectors get $X^*$ as input. A more formal description of the DGP can be found in Appendix \ref{app:details}. Further, throughout we split training data of size $n$ into $n_{train}= \frac{2n}{3}$ for training of all estimators and $n_{val}=\frac{n}{3}$ to be used by the selection criteria, and use an independent test-set of size $n_{test}=500$ to evaluate a criterion by the test-set PEHE of its selected best model. We use $n=1000+500$ as a default, but also consider the effect of having more ($n=2000+1000$) or less ($n=500+250$) data available in Appendix \ref{app:ss}.  Every experiment is repeated for 20 random seeds, across which we report means and standard errors (SEs). Finally, we present additional results using the standard ACIC2016 and IHDP benchmarks in Appendix \ref{app:moredata}, where we also outline why we believe they allow for less interesting analyses -- highlighting further that careful design of DGPs was important to gain interesting insights.}
\begin{figure*}[!t]
	\centering
       \subfigure
       {\includegraphics[width=0.99\textwidth]{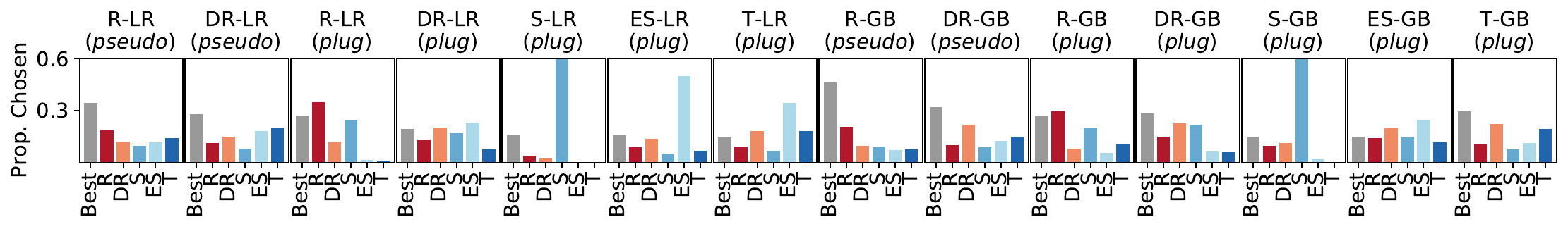}}\vspace{-.3cm}
    \subfigure
    {\includegraphics[width=0.99\textwidth]{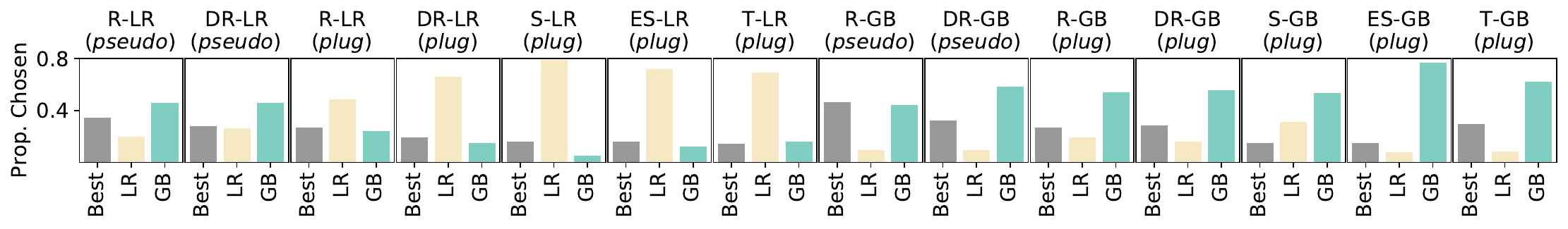}}\vspace{-.5cm}
	\caption{\textbf{Investigating the presence of congeniality bias between selection criteria and estimators: Estimation strategies (top) and ML methods (bottom).} \small Measuring the proportion of times the true best estimator is chosen (left-most bar in each plot), as well as what kind of estimator is chosen when the true best estimator is \textit{not} chosen, across all settings considered. }\label{fig:congstrat}\vspace{-.5cm} 
\end{figure*}
\subsection{Step 2: Examine performance of candidate estimators}
\squish{Here, we briefly examine the performance of the underlying learners themselves in Fig. \ref{fig:learners}. We include an oracle that is trained directly on the (usually unknown) $\tau(X_i)$ to provide a lower bound on error due to misspecification (in particular, this highlights that in setup A and C, LRs cannot capture the CATE well while GBs could). We see that there are indeed interesting performance differences across learners and settings, meaning that no single learner, estimation strategy or ML method consistently dominates all others. As expected, R- and DR-learner show good performance especially when CATE is relatively simple. T- and ES-learners perform worst for $\rho=0$, but relatively better for $\rho$ large; the opposite is true for S-learners. Note that the performance of the LR S-learner is particularly poor for $\rho>0$ because it can only learn a constant treatment effect (i.e. it is severely misspecified for  $\rho>0$). In Appendix \ref{app:otherprop}, we additionally consider imbalanced treatment group sizes and find that this worsens mainly the performance of the R-learner. }

\subsection{Step 3: Towards understanding the performance of different selection criteria}
We are now ready to examine the performance of the different model selection criteria:  In Fig. \ref{fig:overallperf} we present results of different criteria choosing between all 10 learner-method combinations. For legibility, we report performance in terms of $\Delta \text{PEHE}_{fact}$, the difference between the test-PEHE of any given selector and the factual choice (lower is better, negative means better than factual). 

\squish{\textbf{Q1: General performance trends.} In Fig. \ref{fig:overallperf}, we observe that across all selectors, most performance differences and gains relative to factual performance are observed when i) treatment is randomized \textit{and} the treatment effect is complex ($\rho$ large in A\&B), ii) there is confounding \textit{and} the treatment effect is simple ($\rho$ small in C\&D) and iii) LRs are misspecified for the POs but not CATE (B\& D). We observe that the performance of the plug-in criteria often follows the performance of their underlying methods (i.e. comparing the trends in Figs. \ref{fig:learners} and \ref{fig:overallperf}). Further, the plug-in criteria based on ES- and T-learner generally perform the worst, especially when implemented using LRs. In setups C \& D, the plug-in surrogate based on the R-learner works well when CATE complexity is low, but not when it is high -- mimicking the pattern of the underlying estimator observed in the previous section. The pseudo-outcome criteria appear to perform best overall, with the R-pseudo-outcome often performing most similarly to the oracle selector. Note also that while the plug-in criteria based on indirect learners deteriorate substantially when misspecified (LR in Setups A \& C), the pseudo-outcome criteria still perform well even when implemented using a misspecified model (LR).}
\begin{figure*}[!t]
	\centering
       \subfigure[CATE Estimation Performance]
       {\includegraphics[width=0.49\textwidth]{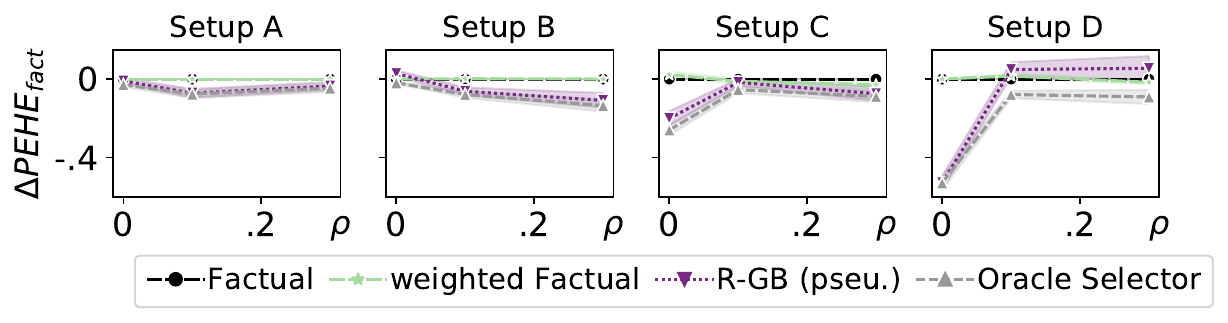}}
    \subfigure[Potential Outcome Estimation Performance]
    {\includegraphics[width=0.49\textwidth]{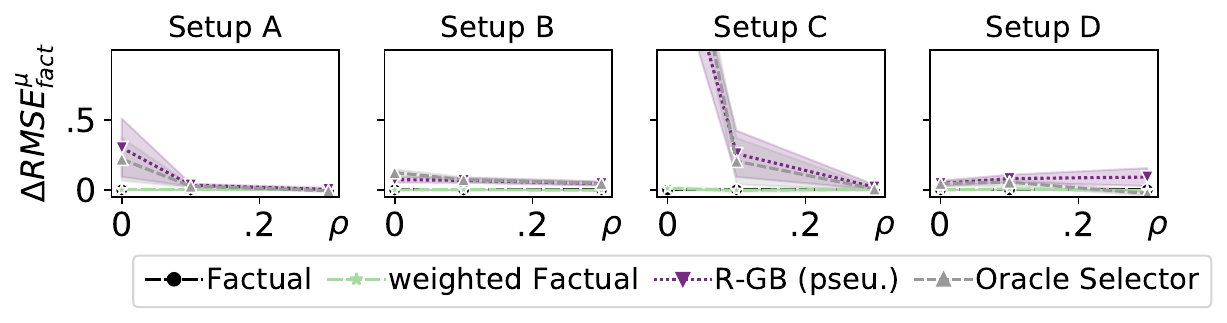}}\vspace{-.2cm}
	\caption{\textbf{Relative performance of different selection criteria when choosing between indirect learners only. } \small  Left: Plotting $\Delta \text{PEHE}_{fact}$, the difference between the test-PEHE of the factual choice and the model selected by any given selection criterion. Right:   Plotting 
 $\Delta \text{RMSE}^\mu_{fact}$, the difference between the average RMSE of estimating the potential outcomes using the factual choice and the model selected by any given selection criterion. (In both, lower is better and negative means better than factual.)}\label{fig:fact}   
 \vspace{-.25cm}
\end{figure*}
{\textbf{Q2. Congeniality bias.} Next we examine whether there is evidence for congeniality bias -- i.e. whether plug-in or pseudo-outcome surrogate criteria appear to inherently favor estimators with similar inductive biases as the strategy used to provide a validation target. We propose to measure this by calculating the proportion of times an (i) estimation strategy or (ii) underlying ML method is selected by a validation criterion  \textit{whenever it does not identify the best estimator} (intuitively, we make this distinction because whenever an estimator is the oracle choice, selecting it should not be considered biased).\footnote{\squish{More formally, we investigate how often a selector $\mathcal{E}_j$ chooses the best estimator $\hat{\tau}^*$ that minimizes the true (unobservable) PEHE $\mathcal{E}^{oracle}_\tau$ and, if not this top choice $\hat{\tau}^*$, which other type of estimator is chosen. That is, in Fig. \ref{fig:congstrat}, we measure using the leftmost bar labeled `best' how often the correct best estimator is chosen --i.e. $\hat{P}(\arg\min_{\hat{\tau}_k} \mathcal{E}_j(\hat{\tau}_k)\!=\!\hat{\tau}^*)=1\!-\!\alpha_{\mathcal{E}_j}$, where $\alpha_{\mathcal{E}_j}$ is the error-rate of the selector -- , and other bars measure the proportion of times any specific type of estimator $\hat{\tau}_l\neq \hat{\tau}^*$ is chosen whenever $\mathcal{E}_j$ \textit{does not make the right choice}. If some type of estimator is chosen disproportionately often -- i.e. if  $\hat{P}(\arg\min_{\hat{\tau}_k} \mathcal{E}_j(\hat{\tau}_k)=\hat{\tau}_l|\hat{\tau}_l\neq \hat{\tau}^*)=1-\alpha_{\mathcal{E}_j}\!\!>>\!\frac{\alpha_{\mathcal{E}_j}}{|\mathcal{T}|}$ -- we consider this evidence that the selector $\mathcal{E}_j$ may be biased towards choosing estimators of the type $\hat{\tau}_l$.} } In Appendix \ref{app:cong}, we present a similar plot without making this distinction.}

In Fig. \ref{fig:congstrat} (top) we investigate congeniality between selection criteria and estimator strategy (i.e. R-, DR-, S-, ES- or T-learner, implemented using \textit{either} ML method), pooled across all settings of Fig. \ref{fig:overallperf}.  We observe that there is clear evidence for congeniality bias between some of the \textit{plug-in} criteria and their corresponding learning strategy; this is most pronounced for the criteria relying on indirect learners, the plug-in S-learner and ES-learner in particular. The plug-in criterion based on the R-learner also clearly suffers from this, while the DR-plug-in criterion exhibits less of this behavior.  The pseudo-outcome criteria overall display less pronounced preference for their own strategy, with the LR-implementations of pseudo-outcome R- and DR-criteria giving least evidence for such congeniality bias overall. 

\squish{In Fig. \ref{fig:congstrat} (bottom) we then investigate congeniality bias between selection criteria and estimator method (i.e. LR or GB, used with \textit{any} estimation strategy). Also here we observe clear evidence for congeniality biases in almost all criteria: LR- (GB-)based criteria appear to prefer learners implemented using LR (GB). The only exception appears to be the LR-pseudo-outcome selectors, who actually select GB learners more often; this may partially explain their good relative performance compared to the other LR-based selectors.}

{\textbf{Q3. What do we lose through factual evaluation?} Finally, we consider the question of \textit{why} we see in Fig. \ref{fig:overallperf} that so many selectors using surrogates for the treatment effect can outperform factual selection in some scenarios. One possible explanation would be the exclusion of direct learners from the candidate pool  available to $\mathcal{E}^{fact}_{Y}$-- however, as can be seen in Fig. \ref{fig:learners}, there is often \textit{some} indirect learner that matches performance of a direct learner. Two alternative explanations  we wish to test would be i) the presence of covariate shift due to confounding and ii) the incorrect focus of factual selection on performance in terms of estimating the POs. We can test i) by including an importance weighted factual selector, and ii) by restricting the candidate estimator pool available to \textit{all selectors} to $\mathcal{T}_{indirect}$ (i.e. excluding R- and DR-learner from the estimator pool). 

In Fig. \ref{fig:fact}(left) we observe that i) does not seem to be the case as weighted and unweighted factual selection perform identically (one possible explanation for this is that none of the considered indirect estimators perform an internal covariate shift correction themselves). Considering ii) we do however observe that indeed both oracle and pseudo-outcome selectors appear to select different (better) indirect estimators than $\mathcal{E}^{fact}_{Y}$, also in the absence of covariate shift (more saliently in Setup B). In Fig. \ref{fig:fact}(right), we show that these, in turn, indeed perform worse in terms of estimating the POs themselves, a trade-off that we expected in Sec. \ref{sec:demyst}. }

\section{Conclusion} We studied the CATE model selection problem and focused on building understanding of the (dis)advantages of different model selection strategies -- using factuals, plug-in surrogates or pseudo-outcome surrogates -- that have been used or proposed in recent work. Instead of attempting to declare a global `winner', we empirically investigated success- and failure modes of different strategies -- and in doing so found that there \textit{are} scenarios where factual selection can be appropriate but also scenarios where pseudo-outcome surrogate approaches are likely to perform better (only plug-in surrogate approaches seemed likely to underperform throughout). We hope that some of the insights presented here will give a starting point for practitioners able to identify how the likely characteristics of their own application translate to the scenarios we considered -- for this purpose we include an additional digest of our findings in form of a Q\&A with an imaginary reader in Appendix \ref{app:takeaways}. We also highlighted that there is a complex interplay between selection strategies, candidate estimators and the DGP used for testing -- congeniality bias is likely to arise when the inductive biases of estimators and selection strategies align. By doing so, we also hope to have demonstrated to the community the need to conduct more simulation studies relying on carefully constructed DGPs to allow to disentangle different forces at play in this problem, enabling more nuanced analyses.

{\textbf{Limitations. } Finally, note that we do not claim our results to be complete: to allow for interesting and nuanced insights, we needed to restrict our attention to specific questions and candidate estimators. We believe that there are a plethora of interesting questions to explore in this area, of which we only made an initial selection to serve as a starting point for discussion. It would, for example, be an interesting next step to consider how different criteria fare at selection between other classes of estimators, e.g. the method-specific neural-network-based estimators extending the work of \citet{shalit2017estimating}, or, at a more microscopic level, at hyperparameter-tuning for any specific method.  While our experimental results are limited to answering some of the questions we found most intriguing, we hope that the desiderata for experimental design that we discuss and implementations that we provide will allow future research to easily expand to further questions and associated DGPs. }

{\section*{Acknowledgements} We would like to thank Toon Vanderschueren, the members of the vanderschaar-lab and anonymous reviewers for insightful comments and discussions on earlier drafts of this paper. AC gratefully acknowledges funding from AstraZeneca.}

\newpage
\bibliography{example_paper}
\bibliographystyle{icml2023}

\newpage
\appendix
\onecolumn
\section*{Appendix}
This Appendix is structured as follows: In Appendix \ref{app:takeaways} we provide an additional overview of key takeaways in form of a Q\&A with an imaginary reader. In Appendix \ref{app:lit} we present an additional literature review discussing CATE estimators proposed in the recent ML literature. Appendix \ref{app:details} discusses experimental details, and Appendix \ref{app:res} presents additional results of the empirical study presented in the main text. Appendix \ref{app:moredata} presents results on additional datasets.
\section{Takeaways: A Q\&A discussing the key insights from this paper with an imaginary reader}\label{app:takeaways}
 We studied the CATE model selection problem and focussed on building understanding of the (dis)advantages of different model selection strategies -- using factuals, plug-in surrogates or pseudo-outcome surrogates -- that have been used or proposed in recent work. Instead of attempting to declare a global `winner', we empirically investigated success- and failure modes of different strategies in the hope that some of the insights presented here will give a starting point for practitioners able to identify how the likely characteristics of their own application translate to the scenarios we considered. For this purpose and because our results may be extensive to digest, we present a light discussion of takeaways in form of a Q\&A with an imaginary reader below. 

\textit{Disclaimer: as we emphasized above, our results do not and cannot cover all possible scenarios. We answer the questions below based on our own empirical studies as well as intuition we built throughout the paper, but note that these conclusions are based only on the limited (yet nuanced) scenarios we were able to consider.}
\begin{itemize}
    \item \textit{Q: So what is the best model selection criterion?} A: There are no magic bullets (yet?!).
    \item \textit{Q: Fine. What are good candidates then?} A: This appears to depend on your data, but overall we found pseudo-outcome surrogates and factual selection to perform well in different scenarios.
    \item \textit{Q: Let me tell you something about my data then. I have confounded data. What should I do?} A: We observed in experiments that especially when the treatment effect is simple and data is confounded, pseudo-outcome criteria using the R- or DR-objective perform much better than other criteria. When the treatment effect is a more complex function, factual criteria appear to perform better.
    \item \textit{Q: What if I have unconfounded data from a clinical trial?} A: When data is unconfounded and models are correctly specified, we found that the selection criterion has slightly less influence. Only in the setting (B) where the treatment effect is a much simpler function than the POs we observed some improvements in using other criteria -- in this case both plug-in and pseudo-outcome surrogates performed better as $\rho$ increased. 
    \item \textit{Q: I expect the treatment effect heterogeneity to be relatively less pronounced than heterogeneity in outcomes regardless of treatment (there is much more prognostic rather than predictive information). What should I do?} A: Pseudo-outcome surrogates, especially the R-learner objective, appear to work very well in this scenario. 
    \item \textit{Q: I expect the opposite -- treatment effects are  likely a very complex function of characteristics -- what does that mean for me?} In this case, using any surrogates may introduce more noise than they help (especially when datasets are small and confounded); it may be advisable to rely on factual validation in this case.
    \item \textit{Q: I would prefer to rely on factuals for validation because I trust that the most as it doesn't require me to estimate any additional parameters. What do I lose out on?} A: We observed that using factual evaluation can be worse in some scenarios \textit{both} because it means that you can only evaluate a smaller set of estimators \textit{and} because it targets the wrong objective -- this particularly seems to matter when the POs are very complex relative to CATE.
    \item \textit{Q: What is this ``congeniality bias'' you were referring to?} A: We consider congeniality bias the issue that surrogate validation targets may advantage CATE estimators $\hat{\tau}_k(x)$ that are \textit{structurally similar} to the used surrogate, due to being imbued with similar inductive biases.  The term itself is usually used in the psychology literature to indicate that individuals may have a systematic preference for information consistent with
current  beliefs \cite{hart2009feeling}.
    \item \textit{Q: I see! Which type of selection criteria suffer from this the most?} A: In our experiments we found that plug-in surrogate selection criteria appear to suffer from this more than pseudo-outcome selection criteria.
    \item \textit{Q: This is all very insightful -- but you do not cover an aspect of model selection I would be interested in. Do you have any advice for me to design my own empirical study investigating this question?} We agree that there are many more interesting questions and hope to expand on these in the future! In Section 4.2 we present desiderata for insightful experiment design if you want to design your own -- and we will also release our code in the future, which is set up in such a way to allow plug-and-play with new selection criteria and datasets! 
\end{itemize}

\section{Additional Literature Review}\label{app:lit}
\subsection{Further CATE estimation strategies.}
In the main text we focussed on one of two prominent streams in the CATE estimation literature, namely the one relying on so-called meta-learners that can be easily implemented using \textit{any ML prediction method}, as originally proposed by \citet{kunzel2019metalearners} and later extended by \citet{nie2021quasi, kennedy2020optimal, curth2021nonparametric}. 

Many methods proposed recently within the ML literature, however, belong to a second stream of literature focussed on \textit{adapting  specific} ML methods to the CATE estimation context. This literature has overwhelmingly relied on indirect estimation strategies, proposing models that learn to make unbiased predictions of outcome under each treatment (so that CATE can be estimated as thei difference). Very popular in this literature, as originally proposed by \citet{johansson2016learning, shalit2017estimating} and later extended in e.g. \citet{johansson2018learning, hassanpour2019counterfactual, hassanpour2019learning, assaad2020counterfactual, curth2021inductive, curth2021nonparametric} has been the use of neural networks to learn a shared representation $\Phi(x)$ that is then used by two treatment-specific output heads $h_a$ to estimate $\hat{\mu}_a(x)=h_a(\Phi(x))$. In this framework, covariate shift arising due to confounding is usually addressed by addition of a balancing regularization term penalising the discrepancy in distribution of $\Phi(x)$ between treatment groups \cite{shalit2017estimating} and importance weighting \cite{johansson2018learning, hassanpour2019counterfactual, assaad2020counterfactual}. Other work has investigated the use of Gaussian Processes \cite{alaa2018limits}, GANs \cite{yoon2018ganite}, VAEs \cite{louizos2017causal, wu2021beta} and deep kernel learning \cite{zhang2020learning} for PO estimation.

The statistics and econometrics literatures, on the other hand, next to the meta-learner strategies discussed in the main text, have mainly relied on tree-based methods, most prominently using S-learner style BART \cite{hill2011bayesian} and direct estimators in the form of causal trees \cite{athey2016recursive} and causal forests \cite{wager2018estimation, athey2019generalized, hahn2017bayesian}.

\subsection{Further model selection strategies. } The model selection strategies discussed in the main text, i.e. those proposed and/or studied in \citet{rolling2014model, nie2021quasi, saito2020counterfactual, alaa2019validating, schuler2018comparison, mahajan2022empirical} all use the given observational data to estimate model performance directly, imputing only some nuisance parameters or surrogate targets. A separate strand of literature \cite{schuler2017synth, athey2021using, parikh2022validating}, which we did not consider further here, suggests to instead validate causal inference models by learning a generative model from the observational dataset at hand and use it to simulate multiple test datasets that share some characteristics with the dataset of interest but have \textit{known} treatment effect that can be used to compare treatment effect estimates of candidate estimators. The estimator found to perform best on such generated datasets should then be used on the real data \cite{schuler2017synth, athey2021using, parikh2022validating}. Finally, other interesting model selection problems exist in the treatment effect estimation context, e.g. the question of how to choose the best first stage \textit{nuisance estimators} for multi-stage treatment effect estimators considered in \citet{cui2019selective}.

\section{Experimental Details}\label{app:details}
\subsection{Implementation details} All code is written in python in sklearn-style to allow for modularity and ease of reuse. All code is available at \texttt{\url{https://github.com/AliciaCurth/CATESelection}} as well as the vanderschaar-lab repository \texttt{\url{https://github.com/vanderschaarlab}}. 

\paragraph{Underlying ML methods}
\begin{itemize}
\item For linear regressions (LR) with ridge (l2) penalty, we use \texttt{RidgeCV} as implemented in \texttt{sklearn} \cite{sklearn_api}; this allows us to automatically implement a sweep over ridge penalties $\lambda \in \{10^{-4}, 10^{-3}, 10^{-2}, 10^{-1}, 1, 10, 10^2, 10^3, 10^4\}$ whenever LR is used as a subroutine in any meta-learner (to estimate $\mu(x)$,  $\mu_0(x)$, $\mu_1(x)$, or  $\tau(x)$).
\item For extreme gradient boosted trees (GB), we use the \texttt{XGBRegressor} as implemented in the \texttt{xgboost} python package  \cite{chen2016xgboost}. Whenever GB is used as a subroutine in any meta-learner, we use \texttt{sklearn}'s 5-fold \texttt{GridSearchCV} to perform a sweep over all combinations of \texttt{learning\_rate} $\in \{.1, .3\}$, \texttt{max\_depth} $\in \{1, 3, 6\}$ and \texttt{n\_estimators} $\in \{20, 100\}$, which are hyperparameters that we observed to have an impact on all learners.
\item Finally, for logistic regressions that used to estimate propensity scores $\pi(x)$ we use the \texttt{sklearn} implementation \texttt{LogisticRegressionCV}, allowing us to sweep over regularization parameters $\lambda \in \{10^{-5}, 10^{-3},10^{-2},10^{-1},1\}$ whenever it is used. 
\end{itemize}  
Note that we reoptimize hyperparameters as part of all subroutines $\mathcal{M}$, which means they are chosen anew for every meta-learner in every seed of every experiment.

\paragraph{Meta-learners} Given a ML-method $\mathcal{M}$, implemented as discussed above, we train the different meta-learners as follows:
\begin{itemize}
    \item S-learner: Append $A$ to $X$ to give $X'=(X, A)$. Call $\mathcal{M}\texttt{.fit}(X', Y)$.
    \item Extended S-learner (ES-learner): Append $A$ and $A*X$ to $X$ to give $X'=(X, A*X, A)$. Call $\mathcal{M}\texttt{.fit}(X', Y)$.
    \item T-learner: Separate data by treatment indicator. Call $\mathcal{M}\texttt{.fit}(X[A==a], Y[A==a])$ for each treatment group $a \in \{0, 1\}$
    \item DR-learner: Call T-learner to get estimates of $\mu_a(x)$. Fit propensity estimate by calling $\mathcal{M}\texttt{.fit}(X, A)$. Use these estimates to compute pseudo-outcome $Y_{DR, \tilde{\eta}}$ as specified in the main text. Call $\mathcal{M}\texttt{.fit}(X, Y_{DR, \tilde{\eta}})$. (Note: we also tested using 5-fold cross-fitting as suggested by \citet{kennedy2020optimal} to ensure consistency, but did not find this to improve performance).
    \item R-learner: Fit propensity estimate as in DR-learner. Fit unconditional mean estimate $\mu(x)$ by calling $\mathcal{M}\texttt{.fit}(X, A)$. Compute pseudo-outcome ${Y}_{R, \tilde{\eta}} = \frac{Y_i - \tilde{\mu}(X_i)}{(A_i - \tilde{\pi}(X_i)}$ and weights $\beta_i = (A_i - \tilde{\pi}(X_i))^2$. Call $\mathcal{M}\texttt{.fit}(X, Y_{R, \tilde{\eta}}, \texttt{sample\_weight}=\beta)$. (Note: we also tested using 5-fold cross-fitting as suggested by \citet{nie2021quasi} to ensure consistency, but did not find this to improve performance).
\end{itemize}

\paragraph{Selection criteria} All selection criteria are computed by solely considering validation data. For plug-in surrogate criteria, the strategies discussed above are used on the validation data to compute a plug-in estimate of the treatment effect. 

For the pseudo-outcome surrogate criteria, we perform 5-fold cross-fitting to avoid correlation between nuisance estimates and outcomes; that is we split the validation data into 5 folds, and use only the 4 folds a data-point is not in to impute their pseudo-outcome. 

While not presented in the main text, in Appendix \ref{app:others} we also computed pseudo-outcome surrogates using the PW-pseudo outcome $\textstyle{{Y}_{PW, \tilde{\eta}} = \left(\frac{A}{\hat{\pi}(X)}- \frac{(1-A)}{1-\hat{\pi}(X)}\right) Y}$ and the matching pseudo outcome of \citet{rolling2014model}, computed by finding the nearest neighbor in Euclidean distance. We also computed the \citet{alaa2019validating}'s influence function validation criterion, we amounts to selecting $\hat{\tau}_k(x)$ that minimizes 
\begin{equation}
    Y_{IF} = (1- B) \tilde{\tau}(x)^2 + BY(\tilde{\tau}(x) - \hat{\tau}(x)) - D * (\tilde{\tau}(x) - \hat{\tau}_k(x))^2 + \hat{\tau}_k(x)^2
\end{equation}
with $D=A-\tilde{\pi}(x)$, $B=2A C^{-1}$ and $C=\tilde{\pi}(1-\tilde{\pi})$ and $\tilde{\tau}(x)$ is a T-learner estimate. All nuisance parameters are estimated using 5 fold cross-estimation.

\subsection{Data-generating process (DGP)} We build on the DGP used in \cite{curth2021inductive} for our experiments. The main differences lie in that we a) randomly binarize the data to consider the effect of misspecification, b) consider higher order interactions to make differences between CATE and the POs more salient and c) also induce confounding.

We also use the covariate data from the Collaborative Perinatal Project provided\footnote{This can be retrieved from \url{https://jenniferhill7.wixsite.com/acic-2016/competition}} for the first Atlantic Causal Inference Competition (ACIC2016) \cite{dorie2019automated} and process all covariates according to
the transformations used for the competition\footnote{We use the code at \url{https://github.com/vdorie/aciccomp/blob/master/2016/R/transformInput.R}}. The original dataset has $d=58$ covariates, of which we exclude the $3$ categorical ones. Of the remaining $55$ covariates, $5$ are binary, $27$ are count data and $23$ are continuous. Because we found that the existing binary and count data are very sparse, we instead decided to randomly binarize variables, by choosing a random observed value in each column and keep only the $23$ continuous columns to resulting in a new input dataset $X^*$ used to create a DGP that mimics a decision tree. (Note that, while not used for outcome simulation, all other columns remain part of $X^{input}$ given to estimators and selection criteria, so all estimators have to also learn to distinguish informative from uninformative columns.)

Similar to \citet{curth2021inductive}, we then use the input data in a linear model with interaction terms:
\begin{equation}\label{simeq}
    Y_i = c + \sum^d_{j=1}\beta_j X^*_j + \sum^d_{j, l}\beta_{j,l} X^*_j X^*_l +  \sum^d_{j, l, k}\beta_{j,l, k} X^*_j X^*_l  X^*_k+  \sum^d_{j, l, k}\beta_{j,l, k,m} X^*_j X^*_l  X^*_k  X^*_m+   A_i \sum^d_{j=1} \gamma_j X_j + \epsilon_i
\end{equation}
where $\epsilon_i \sim N(0, .1)$, $\beta_j\sim \mathcal{B}(.3)$ and $\gamma_j\sim \mathcal{B}(\rho)$. We include each variable randomly into one first, second and third-order interaction term, for which we then simulate coefficient $\beta_{\cdot}\sim \mathcal{B}(.2)$. We chose for each coefficient to be binary to avoid large variances in the scale of POs and CATE across different runs of a simulation, such that RMSE remains comparable across runs. 

We then simulate confounding by assigning treatment according to a propensity score $\pi(x)=expit(\xi \mathcal{Z}(X^{input}\beta))$ where $\mathcal{Z}(\cdot)$ denotes standardisation across the simulation data and $\beta$ is the linear coefficient from eq. \ref{simeq}, ensuring that all variables are true confounders. Note also that $X^{input}$ --i.e. the data as observed by estimators and selectors -- enters the propensity score, ensuring that a logistic regression is always correctly specified for estimating the propensity score. We experiment with further settings in Appendix \ref{app:otherprop}.

The three main experimental knobs under consideration are thus CATE complexity $\rho$, confounding strength $\xi$ and estimator \& selector access to input data $X$ versus $X^*$. In our experiments we always vary $\rho \in \{0, .1, .3\}$ and define settings:
\begin{itemize}
    \item A: Continuous input data $X^{input}=X$, no confounding $\xi=0$
    \item B: Binarized input data $X^{input}=X^*$, no confounding $\xi=0$
        \item C: Continuous input data $X^{input}=X$, no confounding $\xi=3$
    \item D: Binarized input data $X^{input}=X^*$, no confounding $\xi=3$
\end{itemize}

\section{Additional Results Using the Main DGP}\label{app:res}

\subsection{Additional selection criteria}\label{app:others}
\begin{figure}[!h]
    \centering
    \includegraphics[width=.99\textwidth]{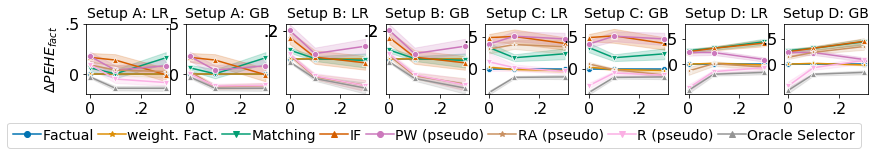}
    \caption{$\Delta \text{PEHE}_{fact}$, the difference between the test-PEHE of any given selection criterion and the factual choice (lower is better, negative means better than factual), for different selection criteria, implemented using linear regressions (LR) and extreme gradient boosting (GB) across 4 different settings, including additional criteria: RA- \& PW-pseudo-outcomes, matching, influence function (IFs) and weighted factual validation (wFactual) not presented in the main text. Here, the complexity of $\tau(x)$ increases in $\rho$. Shaded area indicates one SE.}
    \label{fig:othercrits}
\end{figure}

In Fig. \ref{fig:othercrits} we additionally show performance of matching, two further pseudo-outcomes (PW- and RA-pseudo-outcomes considered in \citet{curth2021nonparametric}), influence function (IFs) and weighted factual validation (wFactual) not presented in the main text. We excluded them in the main text for legibility and because they did not present any improvements over factual selection (if anything they usually performed worse). 

\newpage
\subsection{Additional sample sizes}\label{app:ss}

\begin{figure*}[!h]
	\centering
       \subfigure[$n=500+250$]
       {\includegraphics[width=0.99\textwidth]{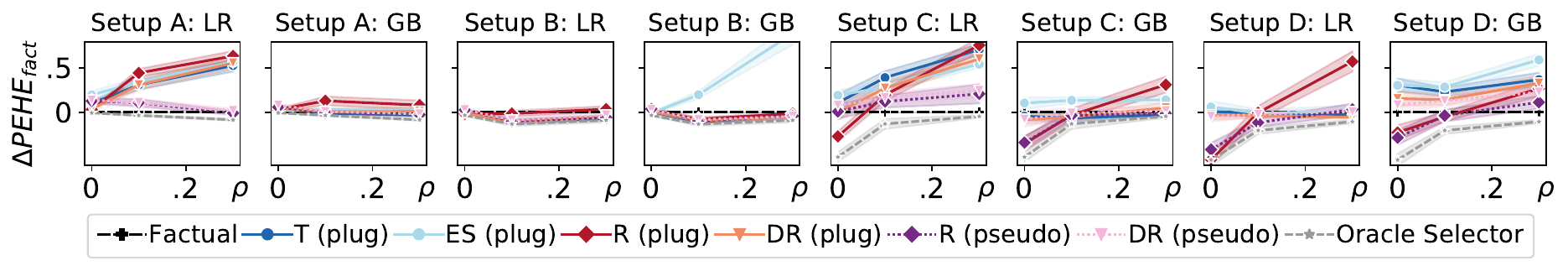}}
              \subfigure[$n=1000+5000$ (reproduced from main text)]
       {\includegraphics[width=0.99\textwidth]{figures/final/main_criteria.pdf}}
       \subfigure[$n=2000+1000$]
       {\includegraphics[width=0.99\textwidth]{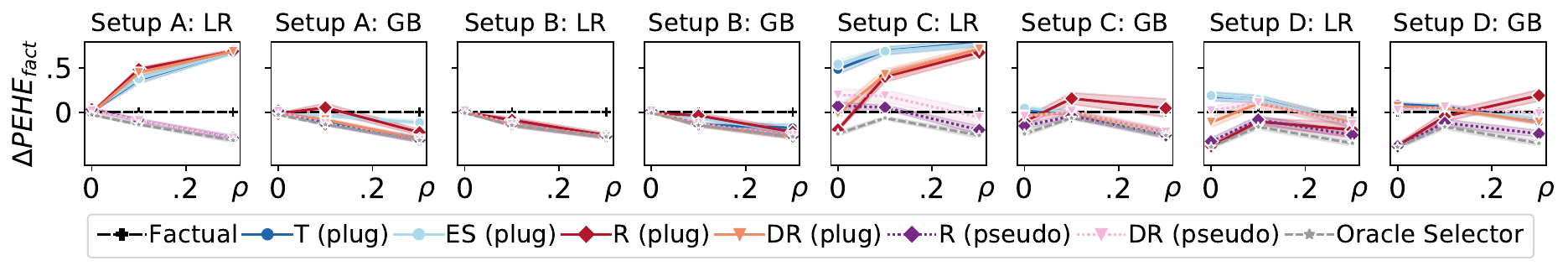}}
	\caption{\textbf{Relative performance of different selection criteria at multiple sample sizes.} \small Plotting $\Delta \text{PEHE}_{fact}$, the difference between the test-PEHE of the factual choice and the model selected by any given selection criterion (lower is better, negative means better than factual) implemented using linear regressions (LR) and extreme gradient boosting (GB).}\label{fig:critsize}  
\end{figure*}
\begin{figure*}[!h]
	\centering
       \subfigure
       {\includegraphics[width=0.99\textwidth]{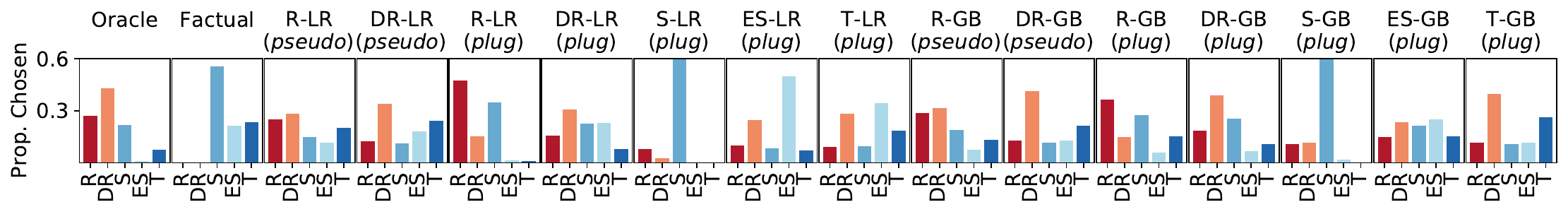}}\vspace{-.3cm}
    \subfigure
    {\includegraphics[width=0.99\textwidth]{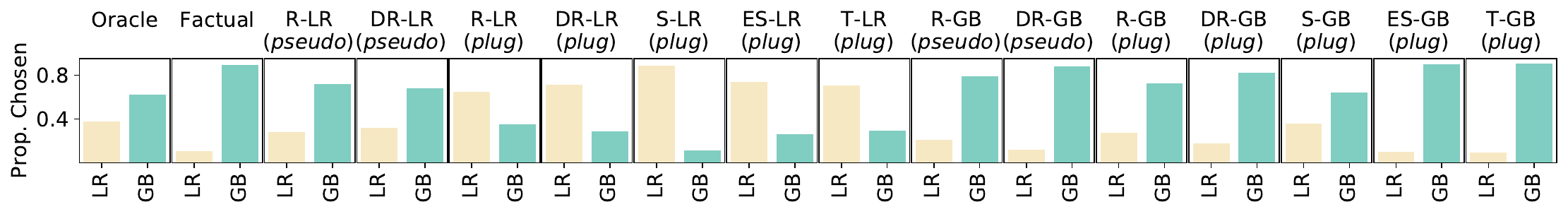}}\vspace{-.5cm}
	\caption{\textbf{Investigating the presence of congeniality bias between selection criteria and estimators: Estimation strategies (top) and ML methods (bottom).} \small Measuring the proportion of times a specific estimator is chosen, across all settings considered in the main text. Note that, different from Fig. \ref{fig:congstrat}, we measure the \textit{absolute} proportion of time any estimator is chosen in this Figure (instead of how often an estimator-type is chosen when a mistake is made as in the main text).}\label{fig:conabs}   
\end{figure*}
\subsection{Additional congeniality plots}\label{app:cong}
In Fig. \ref{fig:conabs}, we present the \textit{absolute} number of times any estimator type is chosen by any selection criterion -- this is different from from Fig. \ref{fig:congstrat} in the main text, which focussed on the types of estimators that are chosen when a selector \textit{makes an error}. While congeniality is much more obvious when we consider only the errors made by selectors, some of the congeniality patterns disucssed in the main text are also clearly reflected in Fig. \ref{fig:conabs}.

\subsection{Additional settings with other propensities}\label{app:otherprop}

\begin{figure*}[!h]
	\centering
              \subfigure[$\pi=0.5$ (reproduced from main text)]
       {\includegraphics[width=0.99\textwidth]{figures/final/main_learner.pdf}}
       \subfigure[$\pi=0.2$]
       {\includegraphics[width=0.99\textwidth]{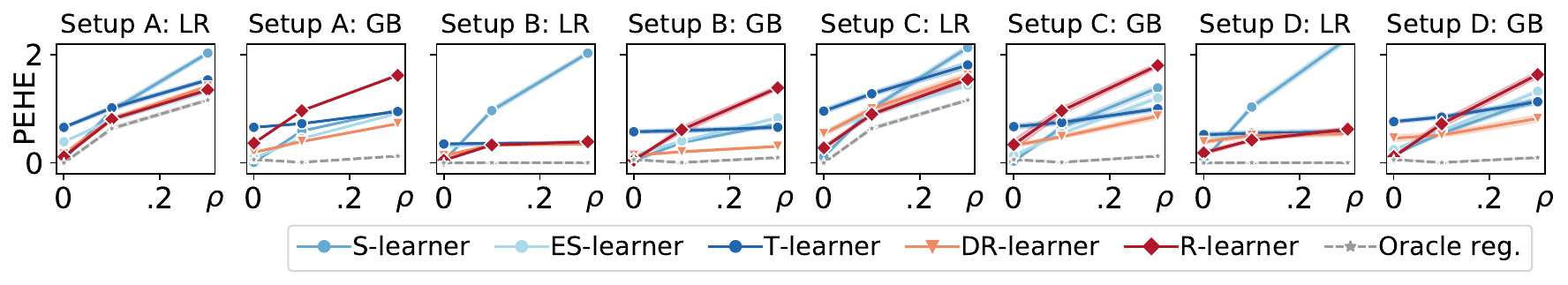}}
	\caption{\textbf{Error in CATE estimation (PEHE) for the different candidate estimators in $\mathcal{T}$, without treatment-group imbalance (top) and with imbalance (bottom).} \small All learners are implemented using linear regressions (LR) and extreme gradient boosting (GB), and considered across 4 different settings where the complexity of $\tau(x)$ increases in $\rho$. Shaded area indicates one SE.}\label{fig:learner_imb}   
\end{figure*}

\begin{figure*}[!h]
	\centering
              \subfigure[$\pi=0.5$ (reproduced from main text)]
       {\includegraphics[width=0.99\textwidth]{figures/final/main_criteria.pdf}}
       \subfigure[$\pi=0.2$]
       {\includegraphics[width=0.99\textwidth]{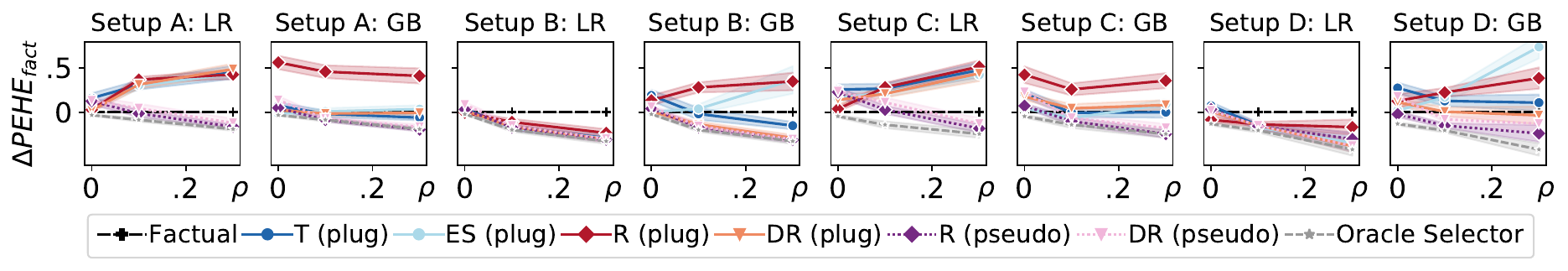}}
	\caption{\textbf{Relative performance of different selection criteria  without treatment-group imbalance (top) and with imbalance (bottom).} \small Plotting $\Delta \text{PEHE}_{fact}$, the difference between the test-PEHE of the factual choice and the model selected by any given selection criterion (lower is better, negative means better than factual) implemented using linear regressions (LR) and extreme gradient boosting (GB). Each criterion gets access to $\mathcal{T}$, i.e. the complete pool of 10 candidate estimators whose performance is shown in Fig. \ref{fig:learner_imb} above. We consider 4 different settings, where the complexity of $\tau(x)$ increases in $\rho$. Shaded area indicates one SE.}\label{fig:crit_imb}   
\end{figure*}

\paragraph{Imbalance.} In Figs. \ref{fig:learner_imb} and \ref{fig:crit_imb} we investigate the effects of adding imbalanced treatment group sizes by re-scaling treatment assignment propensities. Instead of balanced marginal treatment propensity $\pi=0.5$ considered in the main text, we now assign treatment with marginal propensity $\pi=.2$ so that there are substantially more control than treatment units. In Fig. \ref{fig:learner_imb}, in terms of underlying learners, we observe that the most salient difference is that the R-learner now performs relatively worse at large effect heterogeneity. In Fig. \ref{fig:crit_imb}, we find that in the imbalanced setting there is much less improvement over factual selection in the confounded settings with small $\rho$, which is where in the balanced settings there were most performance gains. 

\begin{figure}[!h]
    \centering
    \includegraphics[width=.99\textwidth]{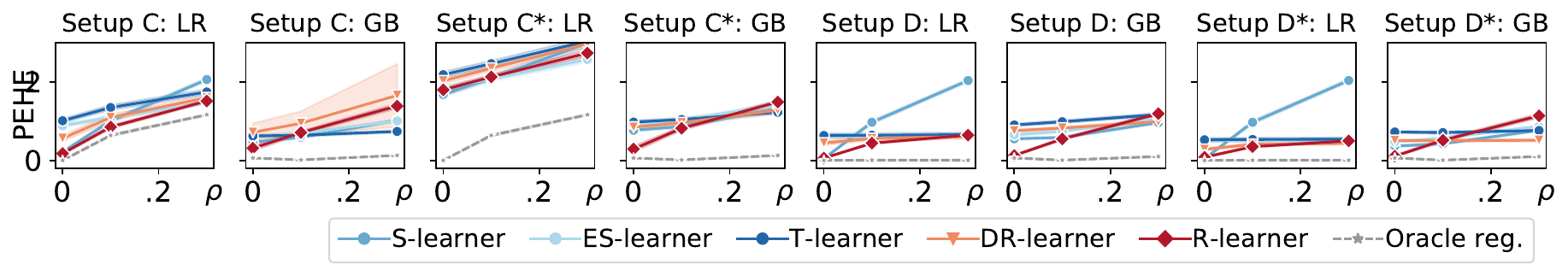}
    \caption{\textbf{Error in CATE estimation (PEHE) for the different candidate estimators in $\mathcal{T}$, for confounded settings with different propensity score specifications.} }
    \label{fig:learner_mis}
\end{figure}

\begin{figure}[!h]
    \centering
    \includegraphics[width=.99\textwidth]{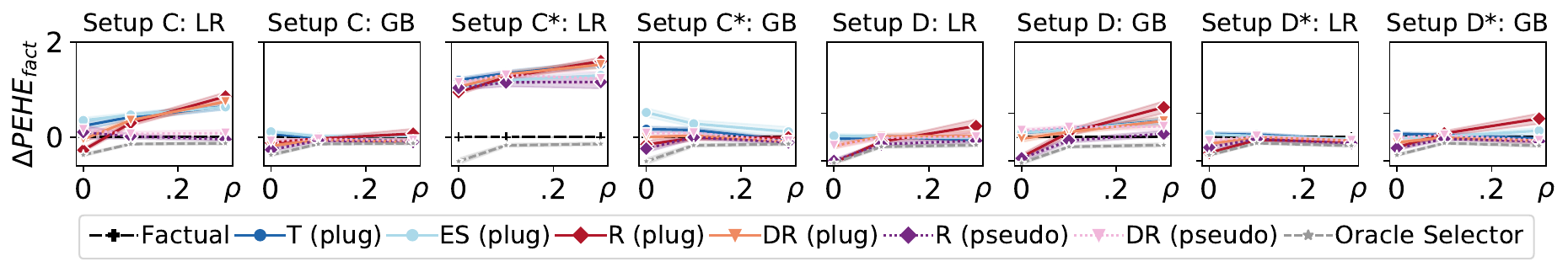}
    \caption{\textbf{Relative performance of different selection criteria, for confounded settings with different propensity score specifications.} }
    \label{fig:crit_mis}
\end{figure}

\paragraph{Other propensity specifications. } In Figs. \ref{fig:learner_mis} and \ref{fig:crit_mis}, we investigate further variation on the used propensity score specification. In particular, note that as discussed in Appendix \ref{app:details}, we ensured that the propensity score is always correctly specified using a logistic regression on $X^{input}$. As a by-product, this means that in Setup C considered in the main text, $\pi$ is a function of $X$ and  $\mu$ is a function of $X^*$, while in Setup D, both $\pi$ and $\mu$ are functions of the binarized (observed) $X^*$ and hence more aligned. Here, we therefore consider two additional setups where $\pi$ is a function of the version of the covariates that is not observed (i.e. $X^*$ for Setup C*, where estimators are given $X$, and $X$ for Setup D*, where estimators are given $X^*$). This could be expected to affect the results in two ways: on the one hand, in the new setup C* and D*, propensity score estimators are misspecified, which could negatively affect estimators and selectors relying on those. On the other hand, note that in the old setup D and the new setup C*, $\pi$ and $\mu$ depend on the same transformation of the covariates $X*$,  while in the old setup C and the new setup D*, $\pi$ and $\mu$ do not depend on the same transformation of the covariates -- in D and C*, propensity scores and outcomes are more aligned, which generally makes estimation harder. 

In  Figs. \ref{fig:learner_mis} and \ref{fig:crit_mis}, we observe that the second effect appears to outweigh the first: especially when using misspecified models (LRs in setups C and C*), for both the candidate estimators and the selection criteria, we observe that aligning $\pi$ and $\mu$ more (i.e. moving from setup C to C*) deteriorates their performance -- regardless of whether a propensity score actually needs to be estimated (e.g. the performance of S- and T-learners, which do not include propensity estimates, also deteriorates). 

\newpage
\section{Additional Datasets: IHDP and ACIC2016}\label{app:moredata}
For completeness, we repeat the experiments conducted in \citet{curth2021really} who highlighted some problems with the use of the ACIC2016 and IHDP datasets in the current CATE estimation literature, in particular the lack of regard for the underlying structural characteristics of these datasets (the setting of experimental knobs they are `biased' towards). With this in mind, we repeat the same experiments from the main text using the setups discussed in \citet{curth2021really}; for details refer to their paper. 

In particular, note that the original IHDP dataset as constructed by \citet{hill2011bayesian} has a very complex treatment effect function, likely advantaging indirect learners. \citet{curth2021really} construct a modified version where the POs remain complex but the treatment effect is simply linear, now possibly advantaging direct learners. Additionally, \citet{curth2021really} study 3 (numbers 2, 26 and 7) of the 77 settings of the ACIC2016 competition \cite{dorie2019automated}: these differ only in the `CATE complexity', where setting 2 has no heterogeneity, 26 has some and 7 is very heterogeneous -- setting similar expectations for relative performance. The IHDP dataset is very small already, with a total size of $747$ so we retain the original train test split and split the training data $2/3$ to $1/3$ for training and validation, as before. The ACIC2016 dataset use the same underlying covariates as our simulations, thus we consider the same three sample size settings. Finally, note that \citet{curth2021really} observed very high variance in absolute RMSE across settings even when using the same estimation strategy. To stabilize results we therefore report relative RMSE ($RMSE(method)/RMSE(baseline)$) where the baseline for all selection criteria remains factual selection, and the baseline for learners on IHDP is T-LR  and T-GB on ACIC2016 (the respective best performing T-learners). 

\paragraph{Comparing underlying learner performance.} We report relative performance of the underlying learners in Table \ref{tab:mod}. We observe that the relative performance of underlying learners is as expected on both datasets and mimics what was shown in \citet{curth2021really}:  direct learners have an advantage on the DGPs with simpler CATE, while T-learners have an advantage on the settings with complex CATE. We also observe that the R-GB learner generally performs worse than in the main text even at low CATE complexity, which may be due to the imbalance in treatment group sizes in all datasets. 
\begin{table}[!h]\label{tab:mod}
\footnotesize
\centering
\setlength\tabcolsep{2pt}
\caption{Relative PEHE of underlying learners on the IHDP and ACIC settings. Averaged across all 100 simulations for IHDP, and across 10 each for ACIC. }
\begin{tabular}{lllllllllllll}
\toprule
Setting              & Oracle-GB & Oracle-LR & S-GB & S-LR & ES-GB & ES-LR & T-GB & T-LR & DR-GB & DR-LR & R-GB & R-LR \\ \midrule
Original IHDP        & 0.58      & 0.78      & 1.58 & 2.58 & 1.54  & 1.58  & 1.14 & 1.00 & 1.25  & 1.11  & 2.60 & 1.65 \\
Modified IHDP        & 0.05      & 0.00      & 0.85 & 0.46 & 1.07  & 1.24  & 1.35 & 1.00 & 0.94  & 0.67  & 4.46 & 1.54 \\
ACIC  2, n=750   & 0.00      & 0.00      & 0.31 & 0.32 & 0.52  & 0.92  & 1.00 & 1.10 & 0.64  & 0.46  & 1.06 & 0.31 \\
ACIC  2, n=1500  & 0.00      & 0.00      & 0.37 & 0.15 & 0.53  & 0.84  & 1.00 & 1.08 & 0.67  & 0.40  & 0.91 & 0.29 \\
ACIC  2, n=3000  & 0.00      & 0.00      & 0.50 & 0.20 & 0.69  & 0.93  & 1.00 & 1.04 & 0.49  & 0.47  & 1.33 & 0.31 \\
ACIC  26, n=750  & 0.59      & 1.07      & 1.12 & 1.61 & 1.04  & 1.26  & 1.00 & 1.28 & 0.94  & 1.26  & 1.24 & 1.27 \\
ACIC  26, n=1500 & 0.58      & 1.32      & 1.20 & 2.03 & 1.08  & 1.57  & 1.00 & 1.47 & 0.91  & 1.46  & 1.22 & 1.47 \\
ACIC  26, n=3000 & 0.58      & 1.55      & 1.20 & 2.41 & 1.03  & 1.74  & 1.00 & 1.71 & 0.95  & 1.70  & 1.40 & 1.72 \\
ACIC  7, n=750   & 1.10      & 1.27      & 1.15 & 1.71 & 1.08  & 1.47  & 1.00 & 1.44 & 0.95  & 1.46  & 1.37 & 1.44 \\
ACIC  7, n=1500  & 0.68      & 1.56      & 1.23 & 2.20 & 1.10  & 1.75  & 1.00 & 1.71 & 1.01  & 1.77  & 1.43 & 1.74 \\
ACIC  7, n=3000  & 0.64      & 1.74      & 1.25 & 2.54 & 1.09  & 1.93  & 1.00 & 1.90 & 0.99  & 1.91  & 1.57 & 1.91\\ \bottomrule
\end{tabular}
\end{table}

\paragraph{Comparing selector performance.} We report relative performance of selectors in Table \ref{tab:sel}. Relative selector performance on IHDP is largely as expected, except that pseudo R- and DR-criterion perform worse than expected on the modified setting, while the plug-in criteria perform better. It is difficult to pinpoint an origin for this, because the IHDP dataset also has i) a much larger control than treated population, ii) limited overlap and iii) very small sample size.  Results on the ACIC datasets are also mixed; here we observe improvements over factual selection mainly for the smallest datasets and when CATE is simple (setting 2). It is possible that this is partially due to the fact that \textit{no} method is able to fit the DGP particularly well with high heterogeneity, seeing as oracle selector performance is only marginally better than factual selection in settings 26 and 7. Note that these ACIC simulations also have limited overlap and imbalances between treatment and control group. We hope that this discussion highlights why we deemed it necessary to construct our own DGPs: Because all these forces are deeply entangled in existing datasets, it is extremely difficult to use them to disambiguate the effects of different factors on performance.
\begin{table}[!h]\label{tab:sel}
\caption{PEHE of model selection metrics relative to factual selection on the IHDP and ACIC settings. Averaged across all 100 simulations for IHDP, and across 10 each for ACIC. }
\footnotesize
\centering
\setlength\tabcolsep{1pt}
\begin{tabular}{lllllllllllllllll}
\toprule
Setting          & Oracle & Factual & S-GB & S-LR & ES-GB & ES-LR & T-GB & T-LR & \begin{tabular}[c]{@{}l@{}}PlugDR\\ GB\end{tabular} & \begin{tabular}[c]{@{}l@{}}PlugDR\\ LR\end{tabular} & \begin{tabular}[c]{@{}l@{}}PlugR\\ GB\end{tabular} & \begin{tabular}[c]{@{}l@{}}PlugR\\ LR\end{tabular} & \begin{tabular}[c]{@{}l@{}}PseuDR\\ GB\end{tabular} & \begin{tabular}[c]{@{}l@{}}PseuDR\\ LR\end{tabular} & \begin{tabular}[c]{@{}l@{}}PseuR\\ GB\end{tabular} & \begin{tabular}[c]{@{}l@{}}PseuR\\ LR\end{tabular} \\ \midrule
Original IHDP    & 0.91   & 1.00    & 1.53 & 2.57 & 1.53  & 1.43  & 1.02 & 1.02 & 1.12                                                & 1.10                                                & 1.81                                               & 1.63                                               & 1.03                                                & 1.03                                                & 1.10                                               & 1.11                                               \\
Modified IHDP    & 0.58   & 1.00    & 0.91 & 0.71 & 0.97  & 1.36  & 0.85 & 0.79 & 0.77                                                & 0.76                                                & 3.32                                               & 1.91                                               & 1.07                                                & 1.01                                                & 1.36                                               & 1.38                                               \\
ACIC 2, n=750    & 0.45   & 1.00    & 2.38 & 6.86 & 2.38  & 5.10  & 3.11 & 3.99 & 3.20                                                & 0.78                                                & 5.57                                               & 2.45                                               & 3.50                                                & 5.80                                                & 4.50                                               & 4.56                                               \\
ACIC  2, n=1500  & 0.35   & 1.00    & 0.54 & 1.13 & 0.98  & 0.90  & 2.27 & 1.83 & 2.08                                                & 1.29                                                & 2.31                                               & 0.51                                               & 2.70                                                & 2.80                                                & 2.16                                               & 2.74                                               \\
ACIC 2, n=3000   & 0.35   & 1.00    & 0.62 & 0.39 & 1.12  & 1.57  & 4.16 & 1.79 & 2.96                                                & 1.04                                                & 3.42                                               & 1.55                                               & 4.83                                                & 4.94                                                & 4.48                                               & 4.21                                               \\
ACIC 26, n=750   & 0.91   & 1.00    & 1.20 & 1.65 & 1.12  & 1.27  & 0.95 & 1.23 & 0.97                                                & 1.26                                                & 1.11                                               & 1.31                                               & 1.00                                                & 0.96                                                & 0.95                                               & 0.96                                               \\
ACIC  26, n=1500 & 0.88   & 1.00    & 1.25 & 2.00 & 1.16  & 1.48  & 0.90 & 1.41 & 0.93                                                & 1.46                                                & 1.30                                               & 1.46                                               & 0.98                                                & 1.00                                                & 0.94                                               & 1.00                                               \\
ACIC 26, n=3000  & 0.86   & 1.00    & 1.25 & 2.46 & 1.08  & 1.79  & 1.00 & 1.78 & 1.01                                                & 1.78                                                & 1.27                                               & 1.77                                               & 1.06                                                & 1.08                                                & 1.12                                               & 1.09                                               \\
ACIC 7, n=750    & 0.85   & 1.00    & 1.34 & 1.65 & 1.12  & 1.38  & 0.90 & 1.35 & 0.90                                                & 1.32                                                & 1.21                                               & 1.46                                               & 0.90                                                & 0.90                                                & 0.88                                               & 0.91                                               \\
ACIC 7, n=1500   & 0.94   & 1.00    & 1.24 & 2.13 & 1.20  & 1.72  & 1.00 & 1.66 & 1.03                                                & 1.66                                                & 1.35                                               & 1.78                                               & 0.99                                                & 1.04                                                & 1.08                                               & 1.07                                               \\
ACIC  7, n=3000  & 0.98   & 1.00    & 1.23 & 2.53 & 1.12  & 1.91  & 0.99 & 1.89 & 1.05                                                & 1.92                                                & 1.25                                               & 1.99                                               & 0.98                                                & 1.04                                                & 1.08                                               & 1.04    \\ \bottomrule                                          
\end{tabular}
\end{table}
\end{document}